\documentclass[final]{clv2025}
\jvol{}
\jnum{}
\jyear{2026}
\jinfo{}
\dochead{Long Paper}
\runningtitle{FOLLIE: French Logical Representations and Automated Reasoning}
\runningauthor{M. Skandalis}
\usepackage[utf8]{inputenc}
\usepackage[T1]{fontenc}
\usepackage{amsmath,amssymb,booktabs,array,tabularx,multirow}
\usepackage{proof}

\usepackage{graphicx,adjustbox,microtype}
\usepackage{tikz}
\usetikzlibrary{arrows.meta,positioning,fit}
\usepackage{enumitem}
\usepackage{xurl}

\newcommand{\unknown}{\ensuremath{\mathsf{unknown}}}

\newcommand{\bo}{[}
\newcommand{\bc}{]}
\newcommand{\bs}{\backslash}
\setlist{itemsep=2pt,topsep=4pt,leftmargin=1.6em}
\hypersetup{pdftitle={FOLLIE: Automated Reasoning over the First Dataset of Sentences with their First-Order Logical Representation for French},pdfauthor={Maximos Skandalis},pdfsubject={French logical semantic parsing and proof-based natural language inference},pdfkeywords={French, semantic parsing, first-order logic, natural language inference, automated reasoning}}
\begin{document}
\title{FOLLIE: Automated Reasoning over the First Dataset of Sentences with their First-Order Logical Representation for French}
\author{Maximos Skandalis}
\affilblock{\affil{LIRMM, CNRS \& University of Montpellier, Montpellier, France}}
\maketitle
\begin{abstract}
FOLLIE (for \underline{F}irst-\underline{O}rder \underline{L}ogic for \underline{L}anguage \underline{I}nference and \underline{E}ntailment for French) connects French sentences, compositional first-order representations
and automated natural language inference (NLI). The first French NLI-ready, NL-to-FOL resource, it contains
19,983 inference records and 36,987 non-empty sentence-slot--formula
mappings from six heterogeneous NLI sources, retaining French lexical
material and source problem structure. Its silver annotations are produced
by a neural--symbolic pipeline, with 111 separately identified LLM-assisted
additions. The construction combines morphosyntactic analysis,
probability-conditioned categorial multitagging, grammatical proof search,
discourse representation and first-order semantic construction. Stage-wise experiments show substantial
gains in derivation coverage; analyses of conjunction distributions,
logical-operator incidence and scope alternation characterise the resulting
resource. We then develop an automatic inference engine whose input-conditioned
background theories address lexical relations, quantification, quantities,
event structure and temporal dependencies. Prover9 supplies proofs for
entailment and contradiction, while Mace4 supports bounded consistency checks.
Three-way accuracies are 65.37\% on all 335 FraCaS-FR problems, 70.33\% on
300 GQNLI-FR problems and 71.35\% on 9,399 eligible SICK-FR records pooled
across splits. Agreement with reference labels among committed decisions is
93.53\%, 100.00\% and 98.66\%, at commitment coverage of 41.49\%, 37.00\%
and 15.05\%, respectively. Worked constructions, axiom schemas and proof
analyses explain these outcomes and identify the contributions of scope,
reference, lexical knowledge and numerical structure. FOLLIE provides a
French experimental basis for studying the representations and assumptions
that connect sentence interpretation to deductive inference. It is publicly available on \url{GitHub} and on \url{Hugging Face}.
\end{abstract}
\section{Introduction}
\label{sec:introduction}

Natural language inference (NLI) connects linguistic interpretation with
consequence: a hypothesis may follow from a premise, contradict it, or remain
compatible without being entailed. Quantifier restrictions, event participants,
lexical relations and temporal reference determine that connection. A logical
representation makes these dependencies available to computation, and a proof
identifies the assumptions from which an inference follows. Bringing the two
together requires a language-to-logic resource that preserves the sentences,
their interpretations and the inference problems in which they occur.

For French, substantial grammatical infrastructure exists
\citep{Moot_2015,Moot2017}, alongside translated and originally authored NLI
resources \citep{conneau-etal-2018-xnli,skandalis-etal-2024-new}. These resources
span controlled semantic problems, image descriptions, application-derived
texts and fact-checking material. Their labels specify an inferential relation;
understanding its linguistic basis also requires the predicate vocabulary,
scope and reference structure of the sentences. Existing logical parsers
construct such meanings, but their analyses have yet to be brought together
in a heterogeneous French sentence--FOL resource organised around those
inference relations.

We introduce FOLLIE, a resource of 19,983 NLI records containing 36,987 non-empty
sentence-slot--formula mappings from six source-corpus families. Its scope is
broad-coverage French inference language, with lexical predicates obtained
from French analysis and source problem identities and labels retained.
To our knowledge, FOLLIE is the first French sentence-to-FOL resource
constructed across heterogeneous NLI corpora that combines French lexical
grounding with preserved source problem structure and labels.
This resource design differs from specialised French geometry formalisation
\citep{tabetgonzalez2021geometry} and from translating English NL--FOL training
inputs into French while keeping the English predicate vocabulary
\citep{vossel2025advancingnaturallanguageformalization}. FOLLIE connects French
sentence-to-logic supervision with the multi-sentence semantic configurations
needed for NLI. The resource, its construction method and its automated
reasoning application form the three contributions of this article.

The construction method combines neural morphosyntactic analysis and
categorial multitagging with GrailLight \citep{Moot2017} for grammatical
proof search. Retaining
probability-conditioned category alternatives raises SICK-FR derivation
coverage from 92.96\% to 99.82\%, with substantial gains also on FraCaS-FR
and GQNLI-FR. The proof terms are interpreted through a French semantic
lexicon as discourse representation structures (DRSs). Our interface
extensions normalise their constructors and derive first-order
representations. Stage-wise evaluation links conversion outcomes to
lexical, grammatical and semantic mechanisms.

The source collections pose different linguistic and inferential demands.
We analyse FOLLIE's coverage, repeated sentence types, annotation
provenance, representation size, logical-operator incidence and scope
alternation. These complementary profiles show how descriptive length and
scope structure vary across corpora: scene descriptions are dominated by
conjunctive event information, whereas less controlled texts combine deeper
Boolean and quantifier dependencies. The resource is silver-standard;
111 separately identified LLM-assisted additions complete selected missing
FraCaS-FR and GQNLI-FR slots. Preserved problem structure supports both
semantic-parser development and investigation of the consequences of its
analyses.

The reasoning contribution is an automatic first-order NLI engine with
input-conditioned lexical, structural, quantitative and temporal background
theories. Prover9 establishes entailment and contradiction through explicit
proof obligations; Mace4 contributes finite model evidence to consistency
checks. Committed labels require successful proofs; \unknown\ is an
operational abstention. No active rule consults a problem identifier or
reference label when generating a verdict.

FraCaS-FR, GQNLI-FR and SICK-FR provide complementary tests of this
architecture, with reference-label agreement above 93\% among commitments
on all three. GQNLI-FR accuracy reaches 70.33\%, compared with 38.33\%
for the evaluated multilingual NLI encoder. Phenomenon-level results and
proof dependencies identify how scope, identity, numerical structure and
background knowledge govern coverage. These experiments evaluate the
representations and rules developed on the three benchmark families.

The linguistic front end extends the French neurosymbolic work of
\citet{skandalis-etal-2025-neurosymbolic}, which coupled neural tagging and
GrailLight with LangPro. Here the principal additions are the FOLLIE
sentence--FOL resource, its structural analysis, the DRS--FOL interface and
the representation-conditioned Prover9/Mace4 engine. LangPro remains a
complementary route in the experimental comparison. 

The article follows the progression from linguistic interpretation to
inferential evaluation. Section~\ref{sec:related} positions FOLLIE among
inference benchmarks, logically annotated corpora and semantic parsing and
reasoning systems. Section~\ref{sec:formalism} establishes the type-logical
and discourse-semantic foundations, including a worked quantified relative
clause (Section~\ref{form:worked-example}) that connects a grammatical
derivation to its first-order interpretation. This background supports the
construction method in Section~\ref{pipe:construction}: we examine the effect
of retaining alternative grammatical categories
(Section~\ref{pipe:multitagging}), develop the DRS-to-FOL interface
(Section~\ref{pipe:translation}), and trace conversion coverage and failures
across its stages (Section~\ref{pipe:coverage}).

Section~\ref{data:resource} characterises the resulting resource through
source provenance, retained sentence--formula coverage, repeated sentences
and model-assisted completion. Its structural analysis
(Section~\ref{data:structures}) combines conjunction distributions with
operator incidence and scope alternation to distinguish the representational
demands of the constituent corpora. Section~\ref{reason:section} then turns
these representations into NLI theories: proof obligations and consistency
guards (Sections~\ref{reason:theory}--\ref{reason:guards}) define when the
system commits to a label, while preprocessing and locally generated axioms
(Sections~\ref{reason:preprocessing}--\ref{reason:axioms}) address the
linguistic information needed for deduction.

Section~\ref{eval:section} evaluates FraCaS-FR, GQNLI-FR and SICK-FR,
relating aggregate precision and coverage to semantic phenomena and axiom
use. Section~\ref{eval:errors} examines representative proof dependencies and
failures, then compares other inference approaches
(Section~\ref{eval:comparison}). Finally, Section~\ref{sec:conclusion}
connects the findings to neural grammatical linking, context-sensitive
lexical knowledge, expert gold-standard annotation and direct French
sentence-to-FOL learning.
\section{Related work}
\label{sec:related}

FOLLIE combines sentence-level logical representations with judgements about
relations between sentences. Its contribution therefore concerns three
connected research questions: what inference resources test, how sentence
meanings are constructed, and how those meanings support automated deduction.

\subsection{Inference benchmarks and linguistic diagnosis}

FraCaS \citep{Pulman1996d} and the Recognising Textual Entailment (RTE)
challenges \citep{10.1007/11736790_9,giampiccolo-etal-2007-third} established
complementary evaluation traditions. The former isolates quantification,
plurals, anaphora, ellipsis, adjectives, comparatives, temporal reference,
verbs and attitudes in hand-built problems, including multi-premise cases.
The latter draws on applications such as information extraction and question
answering, with substantial lexical knowledge and unstated assumptions.
The original entailment/non-entailment distinction was subsequently refined
to distinguish contradiction from other failures of entailment
\citep{de-marneffe-etal-2008-finding}. This annotation history matters for
logical evaluation: non-entailment combines incompatible and compatible
cases, while pragmatic incompatibility can depend on assumptions absent from
a literal first-order interpretation.

SICK \citep{marelli-etal-2014-sick} controls lexical and syntactic variation
in scene descriptions, including negation and active--passive alternation.
SNLI \citep{bowman-etal-2015-large} scales up hypothesis elicitation from
captions; MultiNLI \citep{williams-etal-2018-broad} extends this protocol
across genres. Hypothesis-only artefacts show that classifiers can exploit
collection protocols \citep{poliak-etal-2018-hypothesis}. Diagnostic suites
then distinguish particular dependencies: HANS \citep{mccoy-etal-2019-right}
tests lexical-overlap, subsequence and constituent heuristics; MED
\citep{yanaka-etal-2019-neural} and HELP \citep{yanaka-etal-2019-help}
test polarity-sensitive lexical inference; GQNLI
\citep{cui-etal-2022-generalized-quantifiers} concentrates on generalised
quantifiers and their interactions with negation and scope.

French coverage includes XNLI \citep{conneau-etal-2018-xnli}, designed for
cross-lingual transfer, and FraCaS-FR \citep{amblard-etal-2020-french},
which preserves phenomenon-oriented diagnosis. Additionally, DACCORD \citep{skandalis-etal-2023-daccord,skandalis-etal-2024-new}, and the French versions of RTE3-FR, GQNLI-FR, and SICK-FR \citep{skandalis-etal-2024-new} have expanded the available resources for inference and diagnostic tasks. Their original releases provide
inference judgements without sentence-aligned FOL. FOLLIE adds
representations to heterogeneous French inference sources, allowing the
interpretations and assumptions underlying a predicted relation to be
examined alongside the inherited label.

\subsection{Corpora with logical representations}

Table~\ref{tab:related-resources} compares linguistic source, annotation
unit and formal supervision. These dimensions determine the experiments a
resource supports: query labels supervise reasoning, sentence--formula pairs
supervise interpretation, and consistent symbol interpretations make
separately annotated sentences usable within one theory.

\begin{table}[!tbp]
\centering
\small
\setlength{\tabcolsep}{4pt}
\renewcommand{\arraystretch}{1.12}
\begin{tabular}{@{}>{\raggedright\arraybackslash}p{.24\linewidth}>{\raggedright\arraybackslash}p{.21\linewidth}>{\raggedright\arraybackslash}p{.18\linewidth}>{\raggedright\arraybackslash}p{.28\linewidth}@{}}
\toprule
Resource & Language and source & Annotation unit and scale & Formal target and provenance \\
\midrule
FOLIO \citep{han-etal-2024-folio} & Expert-authored English stories & 1,430 query examples; 487 premise sets & Human-aligned FOL, shared story vocabulary and three-way labels \\
RuleTaker \citep{ijcai2020p0537}; ProofWriter \citep{tafjord-etal-2021-proofwriter} & Generated English facts and rules; paraphrases & Theory--query instances; depth-specific datasets & Logic-program theories and labels; proof graphs in ProofWriter \\
LogicNLI \citep{tian-etal-2021-diagnosing} & Templates with manual linguistic revision & 20K main train/dev/test problems & Unary-predicate formulas, directional derivations and four labels \\
FLD$\times$2 \citep{NEURIPS2024_8678da90} & Proof-first English verbalisation & 110K train/dev/test problems & Logical formulas, derivations and three labels \\
MALLS \citep{yang-etal-2024-harnessing} & LLM-generated English & 28K sentence--formula pairs & 1K human-annotated pairs; 27K filtered automatic targets \\
WillowNLtoFOL \citep{Deveci2026} & Reviewed seeds; LLM-expanded English & 16,014 sentence--formula pairs & FOL; structural checks and selective semantic review \\
FOL-MNLI / FOL-CodeSC \citep{hahn2022formalspecificationsnaturallanguage} & English NLI hypotheses / software descriptions & Approx. 150K / 600K sentence--formula pairs & Automatic C\&C--Boxer FOL; silver targets \\
Parallel Meaning Bank \citep{abzianidze-etal-2017-parallel} & Parallel multilingual texts & Sentence-aligned documents & DRSs, human-verification tiers and derived FOL \\
French geometry \citep{tabetgonzalez2021geometry} & Grammar-generated and manually authored French & $10^6$ generated pairs; approx. 80 manual statements & Geometry-specific syntax trees for coherent first-order implications \\
Multilingual MALLS / Willow \citep{vossel2025advancingnaturallanguageformalization} & French/German machine translations & 35,964 source training pairs & Original FOL retained with English predicate names \\
FOLLIE & Heterogeneous French NLI sources & 36,987 mappings in 19,983 records & Pipeline-derived silver FOL and identified LLM completions; French lexical predicates and source labels \\
\bottomrule
\end{tabular}
\caption{Representative logical resources. Counts retain their stated units; LogicNLI excludes auxiliary diagnostic sets. FOLLIE counts non-empty sentence-slot--formula mappings, including repeated occurrences (Table~\ref{data:composition}).}
\label{tab:related-resources}
\end{table}

RuleTaker \citep{ijcai2020p0537} generates theories of ground facts and
conjunctive rules before verbalising them in English. ProofWriter
\citep{tafjord-etal-2021-proofwriter} adds proof graphs, implication
generation and abduction, with closed- and open-world variants. In the
closed-world setting, failure to derive a fact makes it false; the open-world
setting uses explicit negative information and permits an unknown answer.
Proof depth is measured in the generated rule calculus.
Replacing the rules by classical material implications can additionally
license contraposition, changing the consequences of the generated theory.

Proof-first generation also varies the available derivations. PrOntoQA
\citep{PrOntoQA} verbalises ontology chains supporting repeated modus
ponens. FLD \citep{pmlr-v202-morishita23a} and FLD$\times$2
\citep{NEURIPS2024_8678da90} generate
branching derivation trees and vary deduction rules, distractors and depth,
within propositional and quantified unary-predicate fragments. ProverQA
\citep{qi2025large} combines symbolically generated reasoning skeletons
with LLM-generated story contexts and verbalisation. These methods construct
linguistic examples around a specified deductive structure. LogicNLI
\citep{tian-etal-2021-diagnosing} additionally distinguishes support for a
query, its negation, neither or both. Its fourth label records conflicting
derivations under its directional rules; unrestricted classical consequence
from inconsistent premises would instead entail every query.

FOLIO \citep{han-etal-2024-folio} aligns expert-authored stories with FOL
in a shared story vocabulary. Its Wikipedia-seeded stories and lexicalised
logical templates combine topical variation with controlled deduction;
alignment review makes required background premises explicit. P-FOLIO
\citep{han-etal-2024-p} adds human-written reasoning chains. ProofFOL
\citep{thatikonda2024strategiesimprovingnltofoltranslation} obtains a
larger silver corpus by translating ProofWriter problems and retaining
translations whose Prover9 results match the inherited labels. This tests a
query outcome: different interpretations can agree on that outcome while
disagreeing on other consequences.

For sentence-level translation, MALLS \citep{yang-etal-2024-harnessing}
combines LLM generation, human annotation of a subset and error-driven
filtering. WillowNLtoFOL \citep{Deveci2026} expands reviewed seeds and
adds structurally challenging tests of compositional generalisation.
Independent pairs supply varied predicate vocabularies and logical
structures, while story-level inference additionally requires compatible
symbol use across sentences. Chinese-PL/FOL \citep{lu-etal-2022-parsing}
provides 1,464 Chinese--FOL pairs and 1,263 propositional examples organised
from educational exercises. Its explicit symbol definitions supply part of
the lexical mapping that a parser otherwise has to recover.

Corpus-derived targets preserve another kind of linguistic variation.
The public Text2Log release \citep{9401852} uses \textsc{ccg2lambda} to
generate automatic logical targets from selected English sentences.
FOL-MNLI and FOL-CodeSC \citep{hahn2022formalspecificationsnaturallanguage}
apply C\&C--Boxer to inference hypotheses and software descriptions. Their
silver formulas record the generating parser's analyses of those texts.
The Parallel Meaning Bank \citep{abzianidze-etal-2017-parallel} combines
multilingual DRS annotation with differentiated human-verification status;
discourse referents, scoped operators and semantic roles support richer
comparison than formula counts alone. Annotation review remains relevant
across construction methods: a systematic audit of selected FOLIO and MALLS
data distinguishes ambiguous sentences from incorrect formalisation and
provides corrected targets \citep{brunello2026fixingfoliomallsverified}.

Two French settings further delimit the comparison.
\citet{tabetgonzalez2021geometry} generate syntax trees from a
geometry-specific grammar of coherent first-order implications and
verbalise them in French. Their incidence, equality and perpendicularity
examples concern typed geometric objects; the accompanying manual corpus
also includes anaphoric variants. This supports mathematical formalisation
within a specified domain ontology.
\citet{vossel2025advancingnaturallanguageformalization} translate MALLS
and Willow training inputs into French and German while retaining
English-named predicates; validation and testing remain in English. Their
experiment studies multilingual transfer into a fixed signature.

FOLLIE brings together heterogeneous French NLI texts, derivation-based
silver FOL analyses and identified LLM completions, with French lexical
material and preserved source problem identities and labels. The French
wording is analysed directly, including
when a benchmark was translated from English. Its combination of
sentence-level mappings and premise--hypothesis organisation supports both
French semantic parsing and the study of how separately constructed
interpretations interact in an inference problem.

\subsection{Compositional parsing and neural autoformalisation}

Logical construction involves both selecting a predicate vocabulary and
assembling a formula over it. Recent task analyses distinguish construction
of a labelled signature from logical translation relative to that signature
\citep{brunello2026autoformalization}. These decisions occur in both
compositional parsing and neural autoformalisation.

Categorial derivations constrain how typed lexical meanings combine
\citep{Lambek1958,Moortgat1997,MootRetore2012}. The C\&C
\citep{clark-curran-2007-wide} and EasyCCG
\citep{lewis-steedman-2014-ccg} parsers made CCG-based semantic
construction practical at scale. Boxer \citep{bos-2008-wide} composes
lexical DRS templates over CCG derivations and handles selected anaphoric and
presuppositional dependencies. DRT \citep{KampReyle1993,blackburn:inria-00001208}
separates referent introduction from accessibility conditions; its
first-order translation preserves binding across embedded structures.
Event and thematic-role predicates preserve modification and participant
sharing, affecting which entities and events can be identified across
sentences. Section~\ref{sec:formalism} develops the machinery used here.

The \textsc{ccg2lambda} system \citep{martinez-gomez-etal-2016-ccg2lambda}
combines typed templates along a supplied CCG derivation and reduces the
resulting term. Its initial inference configurations use higher-order
formulas and Coq; subsequent systems employ first-order event semantics or
typed first-order encodings. The grammatical derivation, compositional term
and prover input consequently play distinct roles. LangPro's LLFgen
\citep{abzianidze-2017-langpro} retains lexical and constituent structure
in lambda logical forms for natural-logic reasoning.

For French, TLGbank \citep{Moot_2015} derives type-logical analyses
semi-automatically from the French Treebank. DeepGrail
\citep{moot:tel-03452731} predicts lexical formulas, while the Grail
family \citep{Moot2017} supplies grammatical proof search and
compositional terms. Neural linking and constructive supertagging investigate
how learned predictions can guide formally constrained parsing
\citep{b832b874638349bfa737ef073eabb332,kogkalidis-moortgat-2023-geometry}.
FOLLIE extends this infrastructure through an NLI-oriented semantic and
resource pipeline. A complementary French architecture developed by
\citet{le:tel-03132658} converts dependency analyses into CCG derivations
and adapts Boxer to French lexical and grammatical resources, yielding DRS
and FOL analyses in a different grammatical setting.

LLM-based systems move lexical and structural choices into the generator.
LINC \citep{olausson-etal-2023-linc} translates premises and hypotheses
into Prover9 formulas and aggregates prover-derived labels from sampled
translations. Logic-LM \citep{pan-etal-2023-logic} selects task-appropriate
symbolic solvers and returns execution errors for repair. Executor feedback
identifies malformed expressions or incompatible declarations; an executable
formula can still encode a reversed implication or an unintended scope.
Constructor-based generation addresses the structural interface:
CODE4LOGIC \citep{liu-2025-shot} generates Python calls to logical
constructors, and NL2Logic \citep{putra-etal-2026-nl2logic} recursively
builds an abstract syntax tree before deterministic solver-code generation.

Semantic refinement requires a further connection to the source sentence.
CLOVER \citep{ICLR2025_3e592c57} decomposes translation compositionally
and compares candidates using logical-equivalence classes or separating
interpretations. PrefRAG \citep{zhou-liu-2026-prefrag} instead retrieves
preferred and rejected programs to guide an LLM's repairs. Formal comparison
can establish disagreement between formulas; choosing the intended reading
requires linguistic evidence. French controlled-language reformulation
provides another interface: eXa-LM \citep{frydman2026exalm} rewrites inputs
as French facts and rules before compiling extended Horn clauses for a
Prolog meta-interpreter. Its public tests contain French inputs, source
identifiers and expected answers \citep{frydman2025exalmsoftware}; the
formal programs are computed by the parser, without sentence-aligned FOL
targets in these tests. Its evaluation merges unknown with false under a
closed-world assumption. FOLLIE supplies derivation-based French targets
while retaining the source NLI label inventories.

\subsection{Theorem proving for natural-language inference}

Logical NLI combines semantic interpretation, background knowledge and a
decision procedure. Nutcracker \citep{bos-2014-place} couples
Boxer-derived representations with theorem provers and finite-model
builders. Models provide consistency witnesses or counterexamples,
complementing refutational proofs; input-consistency checks and
knowledge-selection policies govern how these results support a label.
COGEX \citep{tatu-moldovan-2006-logic} combines refutation with lexical
and world-knowledge axioms. Its relaxation mechanism removes constraints
from a failed obligation and scores the resulting derivation, producing an
approximate entailment measure whose interpretation depends on the deleted
information.

Coq-based systems use richer representations and programmable inference.
\citet{mineshima-etal-2015-higher} compose higher-order meanings over CCG
analyses and implement construction-specific tactics. On-demand lexical
axiom injection \citep{martinez-gomez-etal-2017-demand} targets unresolved
subgoals, restricting which lexical relations enter the theory. FraCoq
\citep{chatzikyriakidis-bernardy-2019-wide} uses curated Grammatical
Framework analyses and dynamic semantics for discourse dependencies, with
later extensions to tense and aspect
\citep{bernardy-chatzikyriakidis-2021-applied}. Its reported coverage
characterises these curated semantic analyses and checked proofs; automatic
interpretation of raw text and automatic proof construction require separate
assessment.

LangPro \citep{abzianidze-2015-tableau,abzianidze-2017-langpro} implements
an analytic tableau calculus over lambda logical forms, with rules for
quantifier monotonicity, modification and lexical relations. Its grammatical
front end and lexical resources determine the linguistic analyses available
to the tableau engine. The Dutch adaptation
\citep{Abzianidze_Kogkalidis_2021} changes these interfaces while retaining
the central calculus, illustrating how inference infrastructure can be
transferred across languages.

Typed first-order approaches give another route to semantic coverage.
\citet{Haruta_Mineshima_Bekki_2022} combine CCG analysis, degree and event
representations, arithmetic constraints and lexical abduction with Vampire,
addressing comparatives and further inference benchmarks. TPTP
\citep{Sutcliffe2017TPTP} standardises the resulting proof obligations.
The semantic representation, background axioms and interpretation of solver
outcomes remain components of the NLI system, defining what the prover is
asked to establish and how its answer becomes a linguistic prediction.

\section{Formal background: grammatical proofs and discourse semantics}
\label{sec:formalism}
\label{form:background}

Type-logical proofs license grammatical analyses, lambda terms specify
semantic composition, and discourse representations organise referents and
scoped conditions for first-order interpretation.

\subsection{Type-logical syntax and compositional terms}
\label{form:tlg}

In type-logical grammar (TLG)
\citep{Lambek1958,Moortgat1997,MootRetore2012}, categorial types encode
lexical combinatory behaviour, and parsing derives a designated result type
from sentence resources. In the Lambek convention,
$A/_iB$ consumes a $B$ on its right, while $B\backslash_iA$ consumes one
on its left. Antecedents are nonempty ordered trees whose leaves are
occurrences; repeated formulas remain distinct resources. Their
mode-indexed composition is written $\circ_i$. In $\Gamma\vdash F:A$,
$F$ is the compositional term. Elimination combines disjoint premise resources:
\begin{equation}
\frac{\Gamma\vdash F:A/_iB\quad\Delta\vdash a:B}
     {\Gamma\circ_i\Delta\vdash Fa:A}\;/E
\qquad
\frac{\Delta\vdash a:B\quad\Gamma\vdash F:B\backslash_iA}
     {\Delta\circ_i\Gamma\vdash Fa:A}\;\backslash E.
\label{form:elimination}
\end{equation}
Introduction abstracts over a discharged hypothetical leaf. Mode-specific
structural postulates control exchange, reassociation and constituency.
Combinatory categorial grammar (CCG) \citep{10.5555/332037} assigns
semantic operations to combinatory rules, including forward composition
$F:X/Y,G:Y/Z\mapsto\lambda z.F(Gz):X/Z$.

TLGbank \citep{Moot_2015} reconstructs proofs from the French Treebank and
distinguishes noun, noun-phrase, preposition-specific, and clausal categories.
Its modalised extraction schema is
\begin{equation}
 \frac{X[Y\circ B]\vdash A}
      {X[Y]\vdash A/\Diamond_1\Box_1B}\;\mathrm{Extr},
 \label{form:extraction}
\end{equation}
where $X[\,]$ is a tree context. The discharged argument's position must
be licensed by the $\Diamond_1$-controlled structural postulates within
the mode-1 extraction domain
\citep[Section~2 and Appendix~A]{Moot_2015}.

The Curry--Howard homomorphism abstracts from directionality and structural
modal information to obtain a linear compositional term
\citep{Moortgat1997,Barker2003TLG,MootRetore2012}. Lexical denotations
can duplicate semantic arguments. After lexical substitution, normalisation
uses capture-avoiding beta substitution,
$(\lambda x.F)a\to_\beta F[a/x]$, and eta reduction,
$\lambda x.Fx\to_\eta F$ when $x\notin FV(F)$; $FV$ denotes free variables.

\subsection{DRS construction, accessibility, and first-order interpretation}
\label{form:drt}

A basic discourse representation structure (DRS)
\citep{KampReyle1993,Kamp2011} is $K=[U\mid C]$: $U$ is its finite
referent universe and $C$ its jointly required predications, identities,
negations, disjunctions, and DRS implications. Embedded boxes can use accessible
surrounding referents; conditional consequents can additionally use
antecedent referents. Universes confined to a negation or disjunct remain
local. Properness requires every referent occurrence to have an accessible
introduction or contextual antecedent.

Let $e$ denote individuals, $d$ DRS constructions, and
$q_d=(e\to d)\to d$. In the extensional fragment used below,
$h(n)=e\to d$, $h(np)=q_d$, $h(s)=d$, and
$h(A/_iB)=h(B\backslash_iA)=h(B)\to h(A)$; modalities leave the
semantic type unchanged. The implemented lexicon further refines clauses
as eventuality properties. Merge $\oplus$ combines universes and conditions
before closure, preserving intended sharing; independent referents must be
renamed apart \citep{blackburn:inria-00001208,bos-2008-wide}.

The ordinary DRT-to-FOL recursion is
\citep[Sections~1.5 and~2.2]{KampReyle1993}
\begin{align}
 \mathcal T([u_1,\ldots,u_n\mid C])
   &=\exists u_1\cdots\exists u_n\bigwedge_{c\in C}c^*,
       \label{form:basic-translation}\\
 R(\mathbf t)^*&=R(\mathbf t),\qquad(t=u)^*=(t=u),\nonumber\\
 (\neg K)^*&=\neg\mathcal T(K),\qquad
 (K_1\lor K_2)^*=\mathcal T(K_1)\lor\mathcal T(K_2),\nonumber\\
 ([\mathbf u\mid C]\Rightarrow K)^*
   &=\forall\mathbf u\left(\left(\bigwedge_{c\in C}c^*\right)
                             \to\mathcal T(K)\right).
       \label{form:conditional-translation}
\end{align}
The empty conjunction denotes $\top$. Antecedent referents receive
universal scope over restriction and consequent; positive box referents
receive existential force. For proper basic DRSs with consistent renaming
and matching predicate interpretations, structural induction establishes
satisfaction preservation \citep{KampReyle1993,blackburn:inria-00001208}.
Closure requires an in-scope binder for every variable occurrence.

Presupposition accommodation must retain local binding dependencies
\citep{bos-2003-implementing,blackburn:inria-00001208}. Labelled content
$\ell:K$ can describe a situation without asserting it at the current
evaluation situation \citep{bos-2014-place}; its evaluation index is
separate from the eventuality and participant arguments of verbal
predication.

\subsection{Reference composition for a quantified relative clause}
\label{form:worked-example}

Consider the FraCaS sentence \emph{ITEL entretient tous les ordinateurs
que GFI possède}. A distributive reference reading shows how extraction
and nominal restriction interact with universal scope. Treat \emph{tous les}
as one entry and suppress clause features. The names $I,G$ have category
$np$, the verbs $M,V$ have $(np\backslash s)/np$, the determiner $D$
has $np/n$, the noun $N$ has $n$, and the relative pronoun $R$ has
$(n\backslash n)/(s/\Diamond_1\Box_1np)$. The relative derivation consumes
a hypothetical object $x:np$ and discharges it through extraction:
\begin{equation}
 \frac{G\circ(V\circ x)\vdash VxG:s}
      {G\circ V\vdash\lambda x.VxG:s/\Diamond_1\Box_1np}
      \;\mathrm{Extr}.
 \label{form:itel-extraction}
\end{equation}
The complete term is $M(D(R(\lambda x.VxG)N))I$. Its DRS-valued
lexical interpretation uses fixed name referents $i,g$:
\begin{align*}
 I&=\lambda P.P(i),\qquad G=\lambda P.P(g),\\
 N&=\lambda z.[\,\mid\mathsf{computer}(z)],\\
 V&=\lambda O.\lambda S.S(\lambda u.O(
             \lambda v.[\,\mid\mathsf{possess}(u,v)])),\\
 M&=\lambda O.\lambda S.S(\lambda u.O(
             \lambda v.[\,\mid\mathsf{maintain}(u,v)])),\\
 R&=\lambda Q.\lambda A.\lambda z.
             (A(z)\oplus Q(\lambda P.P(z))),\\
 D&=\lambda A.\lambda B.
             [\,\mid([z\mid\,]\oplus A(z))\Rightarrow B(z)].
\end{align*}
The verbs take their object before their subject, although their relational
atoms display the subject first. Here $I,G:q_d$, $M,V:q_d\to q_d\to d$,
$D:(e\to d)\to q_d$, and $R:(q_d\to d)\to(e\to d)\to(e\to d)$.
The extracted object's semantic type is $q_d$. Applying $R$ to the abstracted clause and
$N$ first constructs
$\lambda z.[\,\mid\mathsf{computer}(z),\mathsf{possess}(g,z)]$.
The determiner places the fresh $z$ and this restriction in a conditional
antecedent; the matrix predicate supplies its consequent:
\begin{align}
 &[\,\mid[z\mid\mathsf{computer}(z),\mathsf{possess}(g,z)]
                         \Rightarrow[\,\mid\mathsf{maintain}(i,z)]],
                   \label{form:itel-drs}\\
 &\forall z\bigl((\mathsf{computer}(z)\land\mathsf{possess}(g,z))
                          \to\mathsf{maintain}(i,z)\bigr).
                   \label{form:itel-fol}
\end{align}
The relative clause restricts the universal. This reference reading abstracts
from tense, plurality and definite-existence presuppositions, allowing an
empty restrictor. The automatic analysis of FraCaS sentence 337, problem 133,
introduces a group referent and a lexical \texttt{tout} condition. The
difference concerns the quantifier's lexical interpretation before standard
DRS-to-FOL translation.

\section{Constructing FOLLIE: from French sentences to first-order representations}
\label{pipe:construction}

We construct FOLLIE with a compositional pipeline that combines neural
category prediction, grammatical proof search, lexical semantic composition,
and first-order translation \citep{rt_thesis}. We retain intermediate annotations and
representations so that tokenisation, argument structure, attachment,
referent introduction, and scope can be inspected at their construction
stages.

For a sentence $s$, the construction has the form
\begin{equation}
 s\longrightarrow (\mathbf{w},\mathbf{p},\boldsymbol{\ell},
                    \mathbf{A}_{\beta})
 \longrightarrow (\pi,L)
 \longrightarrow D
 \longrightarrow \varphi,
 \label{pipe:architecture}
\end{equation}
where $\mathbf{w}$ is the token sequence, $\mathbf{p}$ the part-of-speech
(POS) sequence, $\boldsymbol{\ell}$ the lemmas, and
$\mathbf{A}_{\beta}$ the candidate type-logical categories for the tokens.
GrailLight constructs a derivation $\pi$ and its compositional lambda term $L$.
Lexical substitution and reduction construct a discourse representation $D$;
the last stage compiles its resolved and supported structures into $\varphi$.
Here and below, a ``proof'' at the parsing stage is a grammatical derivation,
not an entailment proof between the sentences of an NLI problem.

We leverage DeepGrail
\citep{moot:tel-03452731} type-logical supertags and POS-tag predictions, and GrailLight \citep{Moot2017}
proof search and lexical semantic composition. We develop the
interfaces needed to process French NLI corpora: maintaining aligned
annotations, extending the Transformer supertagger to retain several candidate
categories, compiling residual discourse constructors, and associating formulas
with their source identifiers. The POS model, chart parser, semantic lexicon,
and ordinary DRS-to-FOL clauses are inherited; our contributions integrate
these components and extend their interfaces for French NLI.

\begin{figure}[!b]
\centering
\begin{tikzpicture}[
 font=\small,
 stage/.style={draw,rounded corners=1pt,align=center,text width=.195\linewidth,
               minimum height=1.2cm,inner sep=4pt},
 >={Latex[length=2mm]}]
\node[stage] (tag) {DeepGrail, Stanza\\Tokens, POS, lemmas\\Candidate types};
\node[stage,right=3mm of tag] (proof) {GrailLight\\Grammatical proof $\pi$\\Lambda term $L$};
\node[stage,right=3mm of proof] (drs) {Semantic lexicon\\Composition and resolution\\Discourse structure $D$};
\node[stage,right=3mm of drs] (fol) {Extended translation script\\Scope and signatures\\Logical form $\varphi$};
\draw[->] (tag) -- (proof);
\draw[->] (proof) -- (drs);
\draw[->] (drs) -- (fol);
\end{tikzpicture}
\caption{Construction stages and retained representations. Multitagging
widens the categories available to proof search; discourse normalisation
fixes binding and accommodation before logical compilation.}
\label{pipe:overview}
\end{figure}
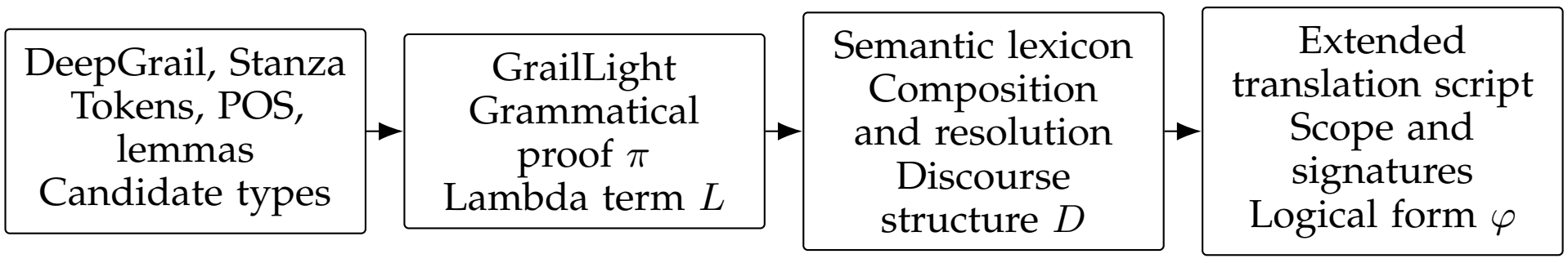

\subsection{A common lexical interface}
\label{pipe:lexical-interface}

We obtain POS tags from the ELMo/bi-LSTM version of DeepGrail
\citep{moot:tel-03452731}, whose tagging component follows
\citet{che-etal-2018-towards}. Its French TreeTagger categories provide
tense and participial distinctions for GrailLight's semantic lexicon.
We check token counts and annotation alignment before parsing: French
elision, contractions, and hyphenation can otherwise shift correspondences
across components.

For lemmatisation, we select Stanza \citep{qi2020stanza} after comparing its lexical outputs
with spaCy \citep{honnibal2020spacy}, using \texttt{fr\_dep\_news\_trf},
and Lefff/MElt \citep{sagot-2010-lefff,denis-sagot-2012-melt} through spaCy. Across 627,997
aligned units, strict pairwise
agreement was 88.39\%, 85.27\%, and 90.14\% for Stanza--spaCy,
Stanza--Lefff/MElt, and spaCy--Lefff/MElt, respectively. These measure output
identity; gold lemmas are unavailable. Differences include decomposing
\emph{du} into \emph{de+le}, retaining the fused form, and normalising
participles to infinitives. Normalisation reduces predicate sparsity but can
obscure adjectival lexicalisation. Lemmas do not have an impact on the proof generation step, but they are directly used in the DRS semantic lexicon and thus, also during the automated reasoning step with the theorem prover at the end.

\subsection{Retaining grammatical alternatives for proof search}
\label{pipe:multitagging}

The Transformer version of DeepGrail predicts a type-logical category for each
token using CamemBERT \citep{martin-etal-2020-camembert} embeddings.
Its categories are learned from TLGbank \citep{Moot_2015} and interpreted
through the proof-theoretic interface defined in
Section~\ref{form:tlg}. We used the already trained model of DeepGrail supertagger from \url{https://gitlab.irit.fr/pnria/global-helper/deepgrail-tagger}.

A one-best supertagger commits to every lexical category before the parser
has checked that the assignments form a derivation. A wrong early decision
can consequently exclude an otherwise available analysis. We extend the
CamemBERT-based model with the relative-probability threshold used in
supertagging by \citet{clark-curran-2004-parsing} and already available in
the earlier ELMo/bi-LSTM DeepGrail \citep{moot:tel-03452731}
implementation. If $p_i(A)$ is the predicted probability of
category $A$ at token $i$, let
\begin{equation}
 A_i^*=\arg\max_A p_i(A),\qquad
 \mathcal A_i(\beta)=
 \{A:p_i(A)\geq\beta p_i(A_i^*)\},\qquad 0<\beta\leq1.
 \label{pipe:threshold}
\end{equation}
The parser receives the retained alternatives and their probabilities.
The $\beta=1$ configuration in our experiments retains one prediction per
token; lowering $\beta$ admits categories that the one-best assignment would
discard. This change extends the decoder's candidate set without retraining
the underlying category classifier.

GrailLight \citep{Moot2017} searches the retained assignments for a complete
derivation. Under the default configuration used here, the category
probabilities rank supertag sequences, while individual parser-rule
applications are unweighted. The parser returns the first derivation found
for the highest-probability sequence licensing a complete proof. Thus,
retaining a correct category can recover grammatical coverage without
resolving attachments among proofs licensed by the same category sequence.
The inherited chart parser also supports rule weighting and preferred
bracketing from a bootstrap parser; those alternatives are not the
configuration evaluated below.

We evaluate $\beta\in\{1,0.1,0.01,0.001,0.0001\}$ across all six
FOLLIE components and their available splits
(Table~\ref{pipe:beta-results}). We hold the lexical annotation pipeline
fixed and measure complete grammatical derivations, ordinary parsing
failures, resource-limit failures, and retained category density. The
GrailLight search-depth limit is 10,000.

\begin{table}[!t]
\centering\footnotesize
\setlength{\tabcolsep}{3.25pt}
\begin{tabular}{@{}llrrrrrrr@{}}
\toprule
Corpus & Split & $N$ & $\beta$ & Parsed & Coverage & Failed & Limits & Types/word\\
\midrule
\multirow{5}{*}{SICK-FR} & \multirow{5}{*}{all} & \multirow{5}{*}{19,680} & 1 & 18,294 & 92.96 & 1,386 & 0 & 1.0000 \\
 & &  & 0.1 & 19,285 & 97.99 & 395 & 0 & 1.0256 \\
 & &  & 0.01 & 19,564 & 99.41 & 116 & 0 & 1.0618 \\
 & &  & 0.001 & 19,619 & 99.69 & 61 & 0 & 1.1431 \\
 & &  & 0.0001 & 19,644 & \textbf{99.82} & 34 & 2 & 1.4157 \\
\addlinespace[2pt]
\multirow{5}{*}{FraCaS-FR} & \multirow{5}{*}{test} & \multirow{5}{*}{882} & 1 & 838 & 95.01 & 44 & 0 & 1.0000 \\
 & &  & 0.1 & 858 & 97.28 & 24 & 0 & 1.0266 \\
 & &  & 0.01 & 869 & 98.53 & 13 & 0 & 1.0819 \\
 & &  & 0.001 & 880 & 99.77 & 2 & 0 & 1.2455 \\
 & &  & 0.0001 & 881 & \textbf{99.89} & 1 & 0 & 1.8624 \\
\addlinespace[2pt]
\multirow{5}{*}{GQNLI-FR} & \multirow{5}{*}{test} & \multirow{5}{*}{703} & 1 & 667 & 94.88 & 36 & 0 & 1.0000 \\
 & &  & 0.1 & 685 & 97.44 & 18 & 0 & 1.0318 \\
 & &  & 0.01 & 688 & 97.87 & 15 & 0 & 1.0562 \\
 & &  & 0.001 & 694 & 98.72 & 9 & 0 & 1.0955 \\
 & &  & 0.0001 & 698 & \textbf{99.29} & 5 & 0 & 1.2444 \\
\addlinespace[2pt]
\multirow{5}{*}{RTE3-FR} & \multirow{5}{*}{test} & \multirow{5}{*}{1,828} & 1 & 1,496 & 81.84 & 332 & 0 & 1.0000 \\
 & &  & 0.1 & 1,707 & 93.38 & 121 & 0 & 1.0528 \\
 & &  & 0.01 & 1,775 & 97.10 & 48 & 5 & 1.1500 \\
 & &  & 0.001 & 1,788 & \textbf{97.81} & 20 & 20 & 1.3974 \\
 & &  & 0.0001 & 1,748 & 95.62 & 12 & 68 & 2.2468 \\
\addlinespace[2pt]
\multirow{5}{*}{RTE3-FR} & \multirow{5}{*}{dev.} & \multirow{5}{*}{1,959} & 1 & 1,593 & 81.32 & 366 & 0 & 1.0000 \\
 & &  & 0.1 & 1,792 & 91.48 & 167 & 0 & 1.0601 \\
 & &  & 0.01 & 1,890 & 96.48 & 57 & 12 & 1.1760 \\
 & &  & 0.001 & 1,899 & \textbf{96.94} & 20 & 40 & 1.4849 \\
 & &  & 0.0001 & 1,837 & 93.77 & 7 & 115 & 2.5643 \\
\addlinespace[2pt]
\multirow{5}{*}{XNLI-FR} & \multirow{5}{*}{test} & \multirow{5}{*}{10,409} & 1 & 8,128 & 78.09 & 2,281 & 0 & 1.0000 \\
 & &  & 0.1 & 9,145 & 87.86 & 1,264 & 0 & 1.0600 \\
 & &  & 0.01 & 9,748 & 93.65 & 661 & 0 & 1.1807 \\
 & &  & 0.001 & 10,045 & \textbf{96.50} & 306 & 58 & 1.5795 \\
 & &  & 0.0001 & -- & -- & -- & -- & 3.0798 \\
\addlinespace[2pt]
\multirow{5}{*}{XNLI-FR} & \multirow{5}{*}{dev.} & \multirow{5}{*}{5,151} & 1 & 3,956 & 76.80 & 1,195 & 0 & 1.0000 \\
 & &  & 0.1 & 4,519 & 87.73 & 632 & 0 & 1.0624 \\
 & &  & 0.01 & 4,824 & 93.65 & 327 & 0 & 1.1913 \\
 & &  & 0.001 & 4,971 & \textbf{96.51} & 144 & 36 & 1.5767 \\
 & &  & 0.0001 & -- & -- & -- & -- & 3.0073 \\
\addlinespace[2pt]
\multirow{5}{*}{DACCORD} & \multirow{5}{*}{all} & \multirow{5}{*}{2,341} & 1 & 1,773 & 75.74 & 568 & 0 & 1.0000 \\
 & &  & 0.1 & 2,053 & 87.70 & 282 & 6 & 1.0684 \\
 & &  & 0.01 & 2,196 & 93.81 & 125 & 20 & 1.1978 \\
 & &  & 0.001 & 2,210 & \textbf{94.40} & 59 & 72 & 1.5470 \\
 & &  & 0.0001 & -- & -- & -- & -- & 2.6925 \\
\bottomrule
\end{tabular}
\caption{Multitagging and grammatical proof coverage for every FOLLIE
component. $N$ counts input sentence units; SICK-FR combines its splits.
Coverage is $100\times\mathrm{Parsed}/N$, rounded to two decimal places;
bold marks the highest evaluated coverage in each group. ``Failed'' excludes
resource-limit failures (``Limits''), so the three outcome counts sum to $N$.
Types/word is the mean number of retained categories. A dash marks an
unavailable proof-search outcome: the final XNLI-FR and DACCORD rows contain
category densities only. Derivation coverage does not establish semantic
correctness.}
\label{pipe:beta-results}
\end{table}

Retaining alternatives recovers derivations on every component, but wider
candidate sets can exhaust bounded proof search. The effect is clear
on RTE3-FR: at the smallest threshold, ordinary failures decline while
resource-limit failures increase enough to reduce coverage. The
XNLI-FR and DACCORD density-only rows quantify the additional lexical
ambiguity without establishing its effect on parsing. Thus $\beta$ controls
a corpus-dependent coverage--search trade-off; category density makes its
computational consequences comparable across runs.

\subsection{Composing discourse representations}
\label{pipe:composition}

The proof term is interpreted through GrailLight's existing French DRT
lexicon, \texttt{big\_french\_drt.pl}.\footnote{The lexicon baseline is GrailLight commit
\texttt{1592b65bacbe66cd921536ed591cb50718b5256c}, 28 May 2026.}
Lexical lookup is indexed by lemma, POS tag, and type-logical formula.
Explicit entries cover semantically sensitive items such as determiners,
pronouns, negation, connectives, and possessives; productive schemata supply
the interpretation of open-class nouns, adjectives, and verbal valency
frames. Verbal interpretation includes eventuality structure, argument
ordering determined by the role lexicon, and temporal constraints selected
by the active tense module. We reuse these lexical resources across the
FOLLIE components.

The lexicon instantiates the DRS-valued composition of
Section~\ref{form:drt}, with eventuality-sensitive verbal meanings and
presuppositional backgrounds. Its intermediate constructor language includes
\texttt{drs}, \texttt{merge}, \texttt{presup}, \texttt{not},
\texttt{bool}, and \texttt{drs\_label}. Lexical substitution and the
inherited capture-avoiding reductions compute these objects; unresolved
\texttt{word(N)} leaves or function-valued applications reveal a failure to
instantiate or normalise the selected analysis. Our extensions compile the
residual discourse constructors after functional reduction.

Some lexical entries consequently represent quantification through predicates
or group conditions instead of explicit universal binders. A group with a
\texttt{tout} condition differs from Equation~\ref{form:itel-fol}; majority
and proportion markers require relationships between groups,
quantities, and individuals. Uninterpreted markers do not supply their
intended generalised-quantifier semantics, a consequential limitation for
GQNLI-FR. Likewise, an ordering symbol or cardinality function requires
supporting arithmetic axioms or an interpreted theory. Universal-binder
counts characterize formula syntax; lexical quantifier conditions require
separate analysis. The DRS identifies the individual, group, or situation
modified by a quantity, supporting correction of its restriction or scope.

\subsection{The DRS-to-FOL interface}
\label{pipe:translation}

Our compiler starts from GrailLight's native \texttt{sem\_utils.pl}, which
already implements lambda reduction, capture-avoiding substitution, ordinary
DRS merge, presupposition-projection heuristics, and the basic DRT-to-FOL
recursion.\footnote{The starting version is GrailLight commit \texttt{4f032988d373d068020bf439dcd3a4b93fe92d07},
26 June 2025.} We extend the interface for the constructors remaining in
French analyses, while keeping semantic construction, discourse reduction,
boundary normalisation, and logical compilation distinct:
\begin{equation}
 \begin{aligned}
 L&\xrightarrow{\text{lexical substitution and }\beta\eta}D_0
  \xrightarrow{\text{DRS reduction}}D_1,\\
 D_1&\xrightarrow{\text{normalisation}}D_2
  \xrightarrow{\mathcal T}\varphi.
 \end{aligned}
 \label{pipe:semantic-stages}
\end{equation}
This ordering matters whenever scope or binding is affected by a residual
constructor. The final recursive translator compiles the selected discourse
reading; it performs no additional presupposition projection or pronoun
resolution.

The ordinary compiler uses Equations~\ref{form:basic-translation}
and~\ref{form:conditional-translation}. Our interface additionally accepts
residual nested \texttt{merge}, generic Boolean operands, and embedded
boxes. Residual merge collects universes and conditions along the
conjunctive structure, deduplicating declarations by referent identity and
retaining intended sharing. Independently introduced referents must already
be standardised apart. Copied descriptions retain accessible enclosing
bindings, while distinct embedded referents remain local; neither negation
nor disjunction makes its local universe globally accessible. A top-level recovery policy adds
declarations omitted upstream. Scope-sensitive free-variable checking then
distinguishes actual closure from the mere occurrence of a same-named binder.

\subsubsection{Presupposition at the attained scope}

We represent an unresolved presuppositional background and its main content
as $\operatorname{presup}(B,M)$. GrailLight's inherited discourse reduction
iterates merge and projection operations. Our extensions strengthen tests
that prevent a presupposition from crossing a binder on which it depends.
This is particularly consequential for DRS implication: a background
depending on an antecedent referent cannot simply be moved outside the
conditional, where that referent is unavailable.

At the boundary of logical compilation, a remaining
$\operatorname{presup}(B,M)$ becomes $B\oplus M$ at the scope it has
already reached. Hence a presupposition that remains inside a negated box
receives local accommodation there. A globally projected reading must have
been selected earlier by discourse reduction. Converting every residual
presupposition to conjunction at the outermost level would make a different
semantic decision. The distinction follows the separation between
construction and resolution in computational DRT
\citep{blackburn:inria-00001208,bos-2003-implementing,bos-2008-wide}.

The labelled-box policy makes a related distinction: non-event referents
needed by a projected background can receive a \emph{de re} reading, whereas
event-dependent backgrounds remain local. We preserve the selected accommodation sites and accessible bindings during
compilation; this policy leaves contextual presupposition resolution to
the preceding discourse analysis.

\subsubsection{Labelled content and its logical signature}

GrailLight uses conditions of the form $\ell:K$, implemented as
\texttt{drs\_label(}$\ell$\texttt{,}$K$\texttt{)}, for subordinate
situations. Such structures occur with attitude verbs and with descriptions
used by subsective modifiers. We extend the ordinary compiler with a
first-order situation encoding for their content, related in purpose to
world-indexed translation in Boxer \citep{bos-2014-place}. Write
$\mathcal I_w$ for the recursive translation inside a labelled description,
evaluated at situation $w$. Its lexical predicates acquire that index:
\begin{align}
 \mathcal T(\ell:K)&=\mathcal I_\ell(K),
          \label{pipe:indexed-box}\\
 \mathcal I_w([U\mid C])&=\exists U\bigwedge_{c\in C}\mathcal I_w(c),
          \nonumber\\
 \mathcal I_w(\ell:K)&=\mathcal I_\ell(K),\nonumber\\
 \mathcal I_w(R(t_1,\ldots,t_n))
   &=R\_\mathrm{event}(w,t_1,\ldots,t_n).
          \label{pipe:indexed-predicate}
\end{align}
Boolean and conditional clauses inside that description thread the index
through their operands; equality and functions retain rigid interpretations.
Outside labelled content, the ordinary clauses preserve the original
predicate signatures. Normalisation declares a label lacking an accessible
declaration in its containing box, while subordinate referents remain local.

The suffix $\_\mathrm{event}$ is an introduced naming convention for these
situation-indexed predicates. Its first argument is an evaluation situation;
the name does not equate that situation with an eventuality introduced inside
$K$. Separate symbols prevent indexed and unindexed versions of a predicate
from acquiring incompatible arities in a prover's signature. The encoding
preserves embedded content without asserting it as an unqualified fact. It
supplies no complete modal theory of belief, tense, or aspect; further
inference depends on the relations licensed between descriptions and their
evaluation situations.

\subsection{A labelled description in the generated representation}
\label{pipe:worked-example}

For the FraCaS sentence \emph{Mickey est un petit animal}, the pipeline
constructs the following analysis. Its reduced
representation has a presuppositional background $B$ and main content $M$:
\begin{align}
 B={}&[y\mid\mathsf{nomm\acute{e}}(y,\mathsf{Mickey}),
          \mathsf{overlaps}(\mathsf{temps}(e),\mathsf{maintenant})],
          \label{pipe:mickey-background}\\
 M={}&[e,\ell,x\mid\mathsf{animal}(x),
          \ell:[\,\mid\mathsf{animal}(x)],
          \mathsf{petit}(\ell),\mathsf{is\_at}(e,x,y)].
          \label{pipe:mickey-main}
\end{align}
Here the temporal referent is shared across the background and main content.
The labelled animal description reuses $x$ from the surrounding box, with no
new subordinate declaration. Accommodation, accessible-variable handling,
and compilation yield, modulo bound-variable renaming:
\begin{equation}
\begin{aligned}
 \exists y\exists e\exists\ell\exists x\bigl(&
 \mathsf{nomm\acute{e}}(y,\mathsf{Mickey})
 \land\mathsf{overlaps}(\mathsf{temps}(e),\mathsf{maintenant})\\
 &\land\mathsf{animal}(x)
 \land\mathsf{animal\_event}(\ell,x)
 \land\mathsf{petit}(\ell)
 \land\mathsf{is\_at}(e,x,y)\bigr).
\end{aligned}
\label{pipe:mickey-export}
\end{equation}
Ordinary predication retains the nominal restriction $\mathsf{animal}(x)$;
the embedded description uses the distinct indexed symbol. The size
condition applies to the description label, not directly to the animal
referent. This records the lexicon's subsective analysis and shows why
matching size predicates alone is insufficient for inference: comparisons
also require a relationship between their descriptions. The copular relation retains its eventuality argument separately from the
description label. This generated reading includes the lexicon's temporal
and definiteness choices, which the distributive reference in
Equation~\ref{form:itel-fol} abstracts away.

\subsection{Conversion coverage and diagnosable failures}
\label{pipe:coverage}

We render representations for NLTK's first-order interface while preserving
source identifiers. Table~\ref{pipe:fol-coverage} gives construction coverage before
assisted completion, using parsed sentence units as denominators. We obtain
36,892 representations without residual lambda or DRS constructors; 36,728
also pass the free-variable check (``Closed''). These checks leave
referential resolution and independent first-order validation open: a bare
\texttt{?} can still mark an unresolved referent.

Construction-stage counts refer to the conversion run; the resource statistics
below count retained aligned slots after assembly and completion. The
36,892 generated representations and 36,876 retained automatic mappings
therefore describe distinct populations, not a single cumulative attrition
curve.

\begin{table}[!tbp]
\centering\small
\begin{tabular}{llrrr}
\hline
Corpus & Split & Parsed units & Generated & Closed\\
\hline
FraCaS-FR & & 881 & 798 & 798 (90.58\%)\\
GQNLI-FR & & 698 & 676 & 676 (96.85\%)\\
SICK-FR & & 19,644 & 19,198 & 19,198 (97.73\%)\\
DACCORD & & 2,210 & 1,695 & 1,677 (75.88\%)\\
\multirow{2}{*}{RTE3-FR} & Development & 1,899 & 1,589 & 1,569 (82.62\%)\\
 & Test & 1,788 & 1,530 & 1,515 (84.73\%)\\
\multirow{2}{*}{XNLI-FR} & Development & 4,971 & 3,769 & 3,737 (75.18\%)\\
 & Test & 10,045 & 7,637 & 7,558 (75.24\%)\\
\hline
Total & & 42,136 & 36,892 & 36,728\\
\hline
\end{tabular}
\caption{Automatic compilation coverage after semantic reduction,
DRS-to-FOL translation, formatting and alignment. Percentages divide
generated representations passing the closure check by parsed units. Counts
precede assisted completion and count sentence occurrences within source
problem structures.}
\label{pipe:fol-coverage}
\end{table}

For SICK-FR, both sentences have a formula in 9,399 of the 9,799 NLI
problems. This problem-level coverage differs from sentence-level coverage:
an isolated converted premise remains useful for sentence-to-FOL work even
if its hypothesis is missing, but it cannot form a complete NLI input.
Likewise, repeated sentences contribute multiple source-aligned units, so
coverage counts are not counts of independent semantic constructions. 

Inspection of failed conversions distinguishes five points of failure.
First, a sentence can lack a complete grammatical derivation. Second, a
derivation can survive lexical substitution with unresolved
\texttt{word(N)} placeholders. Third, the interpreted term can retain
higher-order abstractions or applications outside the target DRS fragment.
Fourth, a linguistically inappropriate category or attachment can propagate
through otherwise valid semantic operations. Fifth, a DRS-form term can be
obtained while translation, formatting, or alignment still fails downstream.
The fourth category can produce a closed formula and is therefore invisible
to a coverage measure based solely on final output.

Representative cases show how the distinctions guide diagnosis. For
\emph{Pitt jura qu'il n'avait rien aperçu de la scène}, a proper-name POS
tag for \emph{jura} selects an atom in place of the required clause-taking
verbal semantics, leaving higher-order arguments unreduced. In FraCaS-FR
353, \emph{Tout rapport a une page de couverture}, the highest-probability
category for \emph{a} is a nominal modifier despite its correct verbal POS
and lemma. The selected proof ends as a noun phrase, and semantic closure
yields a tautological conditional. In GQNLI-FR 50, the selected attachment
places \emph{rouges} on \emph{enfants} instead of \emph{hauts} in
\emph{Six enfants portant des hauts rouges se tiennent debout au sommet
d'une montagne}. The subsequent interpretation can preserve that wrong
attachment without any syntactic error in the formula.

Errors can also originate inside lexical composition. In XNLI-FR test
6595, \emph{En 2004, les États-Unis n'avaient toujours pas tué ou capturé
un seul dirigeant d'Al-Qaïda}, the selected interpretation of \emph{toujours}
applies \emph{pas} to an empty DRS. It introduces $\neg\top$ while the
killing/capturing clause is asserted positively. The inconsistent output is
already determined before the DRS-to-FOL stage. A sound theorem prover
cannot restore the source sentence's meaning from that formula.

Repeated failures include FraCaS-FR comparatives and clitics, SICK-FR
prepositional coordination, and GQNLI-FR comparatives. Length alone does not
explain them: long news sentences can convert while short unsupported
constructions fail. The pipeline provides inspectable construction with
measured derivational and formula-generation coverage; the generated formulas remain
silver analyses requiring targeted semantic correction.

\begin{figure}[t]
\centering
    \begin{minipage}[c]{0.1315\linewidth}
        \begingroup
    \tiny
    \setlength{\baselineskip}{1.15\baselineskip}
        Les données des satellites de la NASA montrent que les calottes glaciaires terrestres de l’Antarctique perdent de la masse depuis 2002.
 \\(\emph{The NASA satellite data show that Earth’s Antarctic ice sheets have been losing mass since 2002.})
            \par
    \endgroup
    \end{minipage}
\hspace{0.15em}$\scriptstyle\rightarrow$
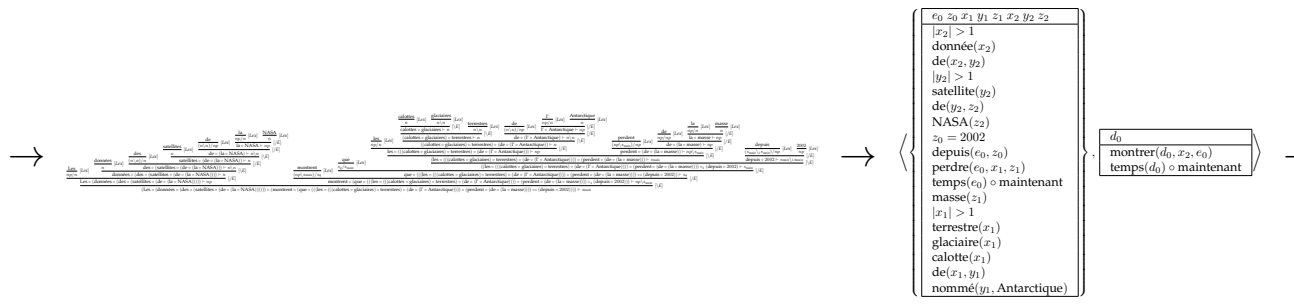
\hspace{0.15em}
\begin{adjustbox}{max width=0.3685\linewidth,valign=c}
$
\infer[\bo\backslash\textit{E}\bc]{(\textrm{Les}\circ_{}(\textrm{données}\circ_{}(\textrm{des}\circ_{}(\textrm{satellites}\circ_{}(\textrm{de}\circ_{}(\textrm{la}\circ_{}\textrm{NASA}))))))\circ_{}(\textrm{montrent}\circ_{}(\textrm{que}\circ_{}(((\textrm{les}\circ_{}(((\textrm{calottes}\circ_{}\textrm{glaciaires})\circ_{}\textrm{terrestres})\circ_{}(\textrm{de}\circ_{}(\textrm{l'}\circ_{}\textrm{Antarctique}))))\circ_{}(\textrm{perdent}\circ_{}(\textrm{de}\circ_{}(\textrm{la}\circ_{}\textrm{masse}))))\circ_{1}(\textrm{depuis}\circ_{}\textrm{$2002$}))))\rule[-.2ex]{0pt}{.9em} \vdash s_{\textit{main}}\rule[-.2ex]{0pt}{.9em}
}{     \infer[\bo/\textit{E}\bc]{\textrm{Les}\circ_{}(\textrm{données}\circ_{}(\textrm{des}\circ_{}(\textrm{satellites}\circ_{}(\textrm{de}\circ_{}(\textrm{la}\circ_{}\textrm{NASA})))))\rule[-.2ex]{0pt}{.9em} \vdash np\rule[-.2ex]{0pt}{.9em}
     }{\infer[\bo\textit{Lex}\bc]{np /_{}n\rule[-.2ex]{0pt}{.9em}
          }{\textrm{Les}\rule[-.2ex]{0pt}{.9em}}
          &
          \infer[\bo\backslash\textit{E}\bc]{\textrm{données}\circ_{}(\textrm{des}\circ_{}(\textrm{satellites}\circ_{}(\textrm{de}\circ_{}(\textrm{la}\circ_{}\textrm{NASA}))))\rule[-.2ex]{0pt}{.9em} \vdash n\rule[-.2ex]{0pt}{.9em}
          }{\infer[\bo\textit{Lex}\bc]{n\rule[-.2ex]{0pt}{.9em}
               }{\textrm{données}\rule[-.2ex]{0pt}{.9em}}
               &
               \infer[\bo/\textit{E}\bc]{\textrm{des}\circ_{}(\textrm{satellites}\circ_{}(\textrm{de}\circ_{}(\textrm{la}\circ_{}\textrm{NASA})))\rule[-.2ex]{0pt}{.9em} \vdash n \bs_{}n\rule[-.2ex]{0pt}{.9em}
               }{\infer[\bo\textit{Lex}\bc]{(n \bs_{}n) /_{}n\rule[-.2ex]{0pt}{.9em}
                    }{\textrm{des}\rule[-.2ex]{0pt}{.9em}}
                    &
                    \infer[\bo\backslash\textit{E}\bc]{\textrm{satellites}\circ_{}(\textrm{de}\circ_{}(\textrm{la}\circ_{}\textrm{NASA}))\rule[-.2ex]{0pt}{.9em} \vdash n\rule[-.2ex]{0pt}{.9em}
                    }{\infer[\bo\textit{Lex}\bc]{n\rule[-.2ex]{0pt}{.9em}
                         }{\textrm{satellites}\rule[-.2ex]{0pt}{.9em}}
                         &
                         \infer[\bo/\textit{E}\bc]{\textrm{de}\circ_{}(\textrm{la}\circ_{}\textrm{NASA})\rule[-.2ex]{0pt}{.9em} \vdash n \bs_{}n\rule[-.2ex]{0pt}{.9em}
                         }{\infer[\bo\textit{Lex}\bc]{(n \bs_{}n) /_{}np\rule[-.2ex]{0pt}{.9em}
                              }{\textrm{de}\rule[-.2ex]{0pt}{.9em}}
                              &
                              \infer[\bo/\textit{E}\bc]{\textrm{la}\circ_{}\textrm{NASA}\rule[-.2ex]{0pt}{.9em} \vdash np\rule[-.2ex]{0pt}{.9em}
                              }{\infer[\bo\textit{Lex}\bc]{np /_{}n\rule[-.2ex]{0pt}{.9em}
                                   }{\textrm{la}\rule[-.2ex]{0pt}{.9em}}
                                   &
\infer[\bo\textit{Lex}\bc]{n\rule[-.2ex]{0pt}{.9em}
                                   }{\textrm{NASA}\rule[-.2ex]{0pt}{.9em}}
                              }
                         }
                    }
               }
          }
     }
     &
     \infer[\bo/\textit{E}\bc]{\textrm{montrent}\circ_{}(\textrm{que}\circ_{}(((\textrm{les}\circ_{}(((\textrm{calottes}\circ_{}\textrm{glaciaires})\circ_{}\textrm{terrestres})\circ_{}(\textrm{de}\circ_{}(\textrm{l'}\circ_{}\textrm{Antarctique}))))\circ_{}(\textrm{perdent}\circ_{}(\textrm{de}\circ_{}(\textrm{la}\circ_{}\textrm{masse}))))\circ_{1}(\textrm{depuis}\circ_{}\textrm{$2002$})))\rule[-.2ex]{0pt}{.9em} \vdash np \bs_{}s_{\textit{main}}\rule[-.2ex]{0pt}{.9em}
     }{\infer[\bo\textit{Lex}\bc]{(np \bs_{}s_{\textit{main}}) /_{}s_{\textit{q}}\rule[-.2ex]{0pt}{.9em}
          }{\textrm{montrent}\rule[-.2ex]{0pt}{.9em}}
          &
          \infer[\bo/\textit{E}\bc]{\textrm{que}\circ_{}(((\textrm{les}\circ_{}(((\textrm{calottes}\circ_{}\textrm{glaciaires})\circ_{}\textrm{terrestres})\circ_{}(\textrm{de}\circ_{}(\textrm{l'}\circ_{}\textrm{Antarctique}))))\circ_{}(\textrm{perdent}\circ_{}(\textrm{de}\circ_{}(\textrm{la}\circ_{}\textrm{masse}))))\circ_{1}(\textrm{depuis}\circ_{}\textrm{$2002$}))\rule[-.2ex]{0pt}{.9em} \vdash s_{\textit{q}}\rule[-.2ex]{0pt}{.9em}
          }{\infer[\bo\textit{Lex}\bc]{s_{\textit{q}} /_{}s_{\textit{main}}\rule[-.2ex]{0pt}{.9em}
               }{\textrm{que}\rule[-.2ex]{0pt}{.9em}}
               &
               \infer[\bo\backslash\textit{E}\bc]{((\textrm{les}\circ_{}(((\textrm{calottes}\circ_{}\textrm{glaciaires})\circ_{}\textrm{terrestres})\circ_{}(\textrm{de}\circ_{}(\textrm{l'}\circ_{}\textrm{Antarctique}))))\circ_{}(\textrm{perdent}\circ_{}(\textrm{de}\circ_{}(\textrm{la}\circ_{}\textrm{masse}))))\circ_{1}(\textrm{depuis}\circ_{}\textrm{$2002$})\rule[-.2ex]{0pt}{.9em} \vdash s_{\textit{main}}\rule[-.2ex]{0pt}{.9em}
               }{                    \infer[\bo\backslash\textit{E}\bc]{(\textrm{les}\circ_{}(((\textrm{calottes}\circ_{}\textrm{glaciaires})\circ_{}\textrm{terrestres})\circ_{}(\textrm{de}\circ_{}(\textrm{l'}\circ_{}\textrm{Antarctique}))))\circ_{}(\textrm{perdent}\circ_{}(\textrm{de}\circ_{}(\textrm{la}\circ_{}\textrm{masse})))\rule[-.2ex]{0pt}{.9em} \vdash s_{\textit{main}}\rule[-.2ex]{0pt}{.9em}
                    }{                         \infer[\bo/\textit{E}\bc]{\textrm{les}\circ_{}(((\textrm{calottes}\circ_{}\textrm{glaciaires})\circ_{}\textrm{terrestres})\circ_{}(\textrm{de}\circ_{}(\textrm{l'}\circ_{}\textrm{Antarctique})))\rule[-.2ex]{0pt}{.9em} \vdash np\rule[-.2ex]{0pt}{.9em}
                         }{\infer[\bo\textit{Lex}\bc]{np /_{}n\rule[-.2ex]{0pt}{.9em}
                              }{\textrm{les}\rule[-.2ex]{0pt}{.9em}}
                              &
                              \infer[\bo\backslash\textit{E}\bc]{((\textrm{calottes}\circ_{}\textrm{glaciaires})\circ_{}\textrm{terrestres})\circ_{}(\textrm{de}\circ_{}(\textrm{l'}\circ_{}\textrm{Antarctique}))\rule[-.2ex]{0pt}{.9em} \vdash n\rule[-.2ex]{0pt}{.9em}
                              }{                                   \infer[\bo\backslash\textit{E}\bc]{(\textrm{calottes}\circ_{}\textrm{glaciaires})\circ_{}\textrm{terrestres}\rule[-.2ex]{0pt}{.9em} \vdash n\rule[-.2ex]{0pt}{.9em}
                                   }{                                        \infer[\bo\backslash\textit{E}\bc]{\textrm{calottes}\circ_{}\textrm{glaciaires}\rule[-.2ex]{0pt}{.9em} \vdash n\rule[-.2ex]{0pt}{.9em}
                                        }{\infer[\bo\textit{Lex}\bc]{n\rule[-.2ex]{0pt}{.9em}
                                             }{\textrm{calottes}\rule[-.2ex]{0pt}{.9em}}
                                             &
\infer[\bo\textit{Lex}\bc]{n \bs_{}n\rule[-.2ex]{0pt}{.9em}
                                             }{\textrm{glaciaires}\rule[-.2ex]{0pt}{.9em}}
                                        }
                                        &
\infer[\bo\textit{Lex}\bc]{n \bs_{}n\rule[-.2ex]{0pt}{.9em}
                                        }{\textrm{terrestres}\rule[-.2ex]{0pt}{.9em}}
                                   }
                                   &
                                   \infer[\bo/\textit{E}\bc]{\textrm{de}\circ_{}(\textrm{l'}\circ_{}\textrm{Antarctique})\rule[-.2ex]{0pt}{.9em} \vdash n \bs_{}n\rule[-.2ex]{0pt}{.9em}
                                   }{\infer[\bo\textit{Lex}\bc]{(n \bs_{}n) /_{}np\rule[-.2ex]{0pt}{.9em}
                                        }{\textrm{de}\rule[-.2ex]{0pt}{.9em}}
                                        &
                                        \infer[\bo/\textit{E}\bc]{\textrm{l'}\circ_{}\textrm{Antarctique}\rule[-.2ex]{0pt}{.9em} \vdash np\rule[-.2ex]{0pt}{.9em}
                                        }{\infer[\bo\textit{Lex}\bc]{np /_{}n\rule[-.2ex]{0pt}{.9em}
                                             }{\textrm{l'}\rule[-.2ex]{0pt}{.9em}}
                                             &
\infer[\bo\textit{Lex}\bc]{n\rule[-.2ex]{0pt}{.9em}
                                             }{\textrm{Antarctique}\rule[-.2ex]{0pt}{.9em}}
                                        }
                                   }
                              }
                         }
                         &
                         \infer[\bo/\textit{E}\bc]{\textrm{perdent}\circ_{}(\textrm{de}\circ_{}(\textrm{la}\circ_{}\textrm{masse}))\rule[-.2ex]{0pt}{.9em} \vdash np \bs_{}s_{\textit{main}}\rule[-.2ex]{0pt}{.9em}
                         }{\infer[\bo\textit{Lex}\bc]{(np \bs_{}s_{\textit{main}}) /_{}np\rule[-.2ex]{0pt}{.9em}
                              }{\textrm{perdent}\rule[-.2ex]{0pt}{.9em}}
                              &
                              \infer[\bo/\textit{E}\bc]{\textrm{de}\circ_{}(\textrm{la}\circ_{}\textrm{masse})\rule[-.2ex]{0pt}{.9em} \vdash np\rule[-.2ex]{0pt}{.9em}
                              }{\infer[\bo\textit{Lex}\bc]{np /_{}np\rule[-.2ex]{0pt}{.9em}
                                   }{\textrm{de}\rule[-.2ex]{0pt}{.9em}}
                                   &
                                   \infer[\bo/\textit{E}\bc]{\textrm{la}\circ_{}\textrm{masse}\rule[-.2ex]{0pt}{.9em} \vdash np\rule[-.2ex]{0pt}{.9em}
                                   }{\infer[\bo\textit{Lex}\bc]{np /_{}n\rule[-.2ex]{0pt}{.9em}
                                        }{\textrm{la}\rule[-.2ex]{0pt}{.9em}}
                                        &
\infer[\bo\textit{Lex}\bc]{n\rule[-.2ex]{0pt}{.9em}
                                        }{\textrm{masse}\rule[-.2ex]{0pt}{.9em}}
                                   }
                              }
                         }
                    }
                    &
                    \infer[\bo/\textit{E}\bc]{\textrm{depuis}\circ_{}\textrm{$2002$}\rule[-.2ex]{0pt}{.9em} \vdash s_{\textit{main}} \bs_{1}s_{\textit{main}}\rule[-.2ex]{0pt}{.9em}
                    }{\infer[\bo\textit{Lex}\bc]{(s_{\textit{main}} \bs_{1}s_{\textit{main}}) /_{}np\rule[-.2ex]{0pt}{.9em}
                         }{\textrm{depuis}\rule[-.2ex]{0pt}{.9em}}
                         &
\infer[\bo\textit{Lex}\bc]{np\rule[-.2ex]{0pt}{.9em}
                         }{\textrm{$2002$}\rule[-.2ex]{0pt}{.9em}}
                    }
               }
          }
     }
}
$
\end{adjustbox}
\hspace{0.15em}$\scriptstyle\rightarrow$\hspace{0.15em}
\begin{adjustbox}{max width=0.1785\linewidth,valign=c}
$
\left\langle
\left\{
\vcenter{\hbox{%
\,\setlength{\tabcolsep}{8pt}%
\begin{tabular}[t]{|l|}
\firsthline
$e_{0}\ z_{0}\ x_{1}\ y_{1}\ z_{1}\ x_{2}\ y_{2}\ z_{2}$ \\ \hline
$|x_{2}|>\textrm{$1$}$ \\[0pt]
$\textrm{donnée}(x_{2})$ \\[0pt]
$\textrm{de}(x_{2},y_{2})$ \\[0pt]
$|y_{2}|>\textrm{$1$}$ \\[0pt]
$\textrm{satellite}(y_{2})$ \\[0pt]
$\textrm{de}(y_{2},z_{2})$ \\[0pt]
$\textrm{NASA}(z_{2})$ \\[0pt]
$z_{0}=\textrm{$2002$}$ \\[0pt]
$\textrm{depuis}(e_{0},z_{0})$ \\[0pt]
$\textrm{perdre}(e_{0},x_{1},z_{1})$ \\[0pt]
$\textrm{temps}(e_{0})\circ\textrm{maintenant}$ \\[0pt]
$\textrm{masse}(z_{1})$ \\[0pt]
$|x_{1}|>\textrm{$1$}$ \\[0pt]
$\textrm{terrestre}(x_{1})$ \\[0pt]
$\textrm{glaciaire}(x_{1})$ \\[0pt]
$\textrm{calotte}(x_{1})$ \\[0pt]
$\textrm{de}(x_{1},y_{1})$ \\[0pt]
$\textrm{nommé}(y_{1},\textrm{Antarctique})$ \\[0pt]
\lasthline
\end{tabular}%
}}
\right\},
\vcenter{\hbox{%
\,\setlength{\tabcolsep}{8pt}%
\begin{tabular}[t]{|l|}
\firsthline
$d_{0}$ \\ \hline
$\textrm{montrer}(d_{0},x_{2},e_{0})$ \\[0pt]
$\textrm{temps}(d_{0})\circ\textrm{maintenant}$ \\[0pt]
\lasthline
\end{tabular}%
}}
\right\rangle
 $
 \end{adjustbox}
\hspace{0.15em}$\scriptstyle\rightarrow$\hspace{0.15em}
    \begin{minipage}[c]{0.1885\linewidth}
    \begingroup
    \tiny
    \ttfamily
    \setlength{\baselineskip}{0.9\baselineskip}
exists a b c d e f g h i.((h = 2002) \& (depuis(i, h) \& (>(num(g), 1) \& (terrestre(g) \& (glaciaire(g) \& (calotte(g) \& (nommé(f, 'Antarctique') \& (de(g, f) \& (masse(e) \& (perdre(i, g, e) \& (overlaps(temps(i), maintenant) \& (>(num(d), 1) \& (>(num(c), 1) \& (satellite(c) \& ('NASA'(b) \& (de(c, b) \& (donnée(d) \& (de(d, c) \& (montrer(a, d, i) \& overlaps(temps(a), maintenant))))))))))))))))))))
    \par
    \endgroup
    \end{minipage}
\\
\vspace{0.5cm}
    \begin{minipage}[c]{0.1315\linewidth}
        \begingroup
    \tiny
    \setlength{\baselineskip}{1.15\baselineskip}
        La police a relâché un suspect qui a tenté de faire exploser une bombe.
 \\(\emph{Police released a suspect who tried to detonate a bomb.})
            \par
    \endgroup
    \end{minipage}
\hspace{0.15em}$\scriptstyle\rightarrow$
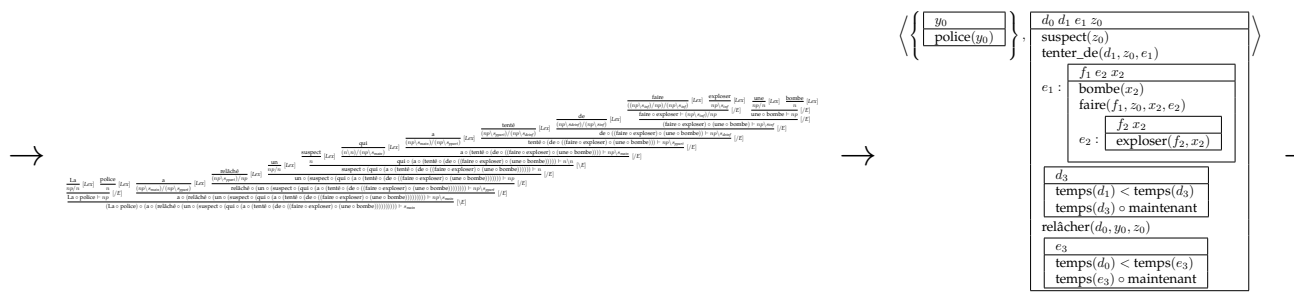
\hspace{0.15em}
\begin{adjustbox}{max width=0.3685\linewidth,valign=c}
$
\infer[\bo\backslash\textit{E}\bc]{(\textrm{La}\circ_{}\textrm{police})\circ_{}(\textrm{a}\circ_{}(\textrm{relâché}\circ_{}(\textrm{un}\circ_{}(\textrm{suspect}\circ_{}(\textrm{qui}\circ_{}(\textrm{a}\circ_{}(\textrm{tenté}\circ_{}(\textrm{de}\circ_{}((\textrm{faire}\circ_{}\textrm{exploser})\circ_{}(\textrm{une}\circ_{}\textrm{bombe}))))))))))\rule[-.2ex]{0pt}{.9em} \vdash s_{\textit{main}}\rule[-.2ex]{0pt}{.9em}
}{     \infer[\bo/\textit{E}\bc]{\textrm{La}\circ_{}\textrm{police}\rule[-.2ex]{0pt}{.9em} \vdash np\rule[-.2ex]{0pt}{.9em}
     }{\infer[\bo\textit{Lex}\bc]{np /_{}n\rule[-.2ex]{0pt}{.9em}
          }{\textrm{La}\rule[-.2ex]{0pt}{.9em}}
          &
\infer[\bo\textit{Lex}\bc]{n\rule[-.2ex]{0pt}{.9em}
          }{\textrm{police}\rule[-.2ex]{0pt}{.9em}}
     }
     &
     \infer[\bo/\textit{E}\bc]{\textrm{a}\circ_{}(\textrm{relâché}\circ_{}(\textrm{un}\circ_{}(\textrm{suspect}\circ_{}(\textrm{qui}\circ_{}(\textrm{a}\circ_{}(\textrm{tenté}\circ_{}(\textrm{de}\circ_{}((\textrm{faire}\circ_{}\textrm{exploser})\circ_{}(\textrm{une}\circ_{}\textrm{bombe})))))))))\rule[-.2ex]{0pt}{.9em} \vdash np \bs_{}s_{\textit{main}}\rule[-.2ex]{0pt}{.9em}
     }{\infer[\bo\textit{Lex}\bc]{(np \bs_{}s_{\textit{main}}) /_{}(np \bs_{}s_{\textit{ppart}})\rule[-.2ex]{0pt}{.9em}
          }{\textrm{a}\rule[-.2ex]{0pt}{.9em}}
          &
          \infer[\bo/\textit{E}\bc]{\textrm{relâché}\circ_{}(\textrm{un}\circ_{}(\textrm{suspect}\circ_{}(\textrm{qui}\circ_{}(\textrm{a}\circ_{}(\textrm{tenté}\circ_{}(\textrm{de}\circ_{}((\textrm{faire}\circ_{}\textrm{exploser})\circ_{}(\textrm{une}\circ_{}\textrm{bombe}))))))))\rule[-.2ex]{0pt}{.9em} \vdash np \bs_{}s_{\textit{ppart}}\rule[-.2ex]{0pt}{.9em}
          }{\infer[\bo\textit{Lex}\bc]{(np \bs_{}s_{\textit{ppart}}) /_{}np\rule[-.2ex]{0pt}{.9em}
               }{\textrm{relâché}\rule[-.2ex]{0pt}{.9em}}
               &
               \infer[\bo/\textit{E}\bc]{\textrm{un}\circ_{}(\textrm{suspect}\circ_{}(\textrm{qui}\circ_{}(\textrm{a}\circ_{}(\textrm{tenté}\circ_{}(\textrm{de}\circ_{}((\textrm{faire}\circ_{}\textrm{exploser})\circ_{}(\textrm{une}\circ_{}\textrm{bombe})))))))\rule[-.2ex]{0pt}{.9em} \vdash np\rule[-.2ex]{0pt}{.9em}
               }{\infer[\bo\textit{Lex}\bc]{np /_{}n\rule[-.2ex]{0pt}{.9em}
                    }{\textrm{un}\rule[-.2ex]{0pt}{.9em}}
                    &
                    \infer[\bo\backslash\textit{E}\bc]{\textrm{suspect}\circ_{}(\textrm{qui}\circ_{}(\textrm{a}\circ_{}(\textrm{tenté}\circ_{}(\textrm{de}\circ_{}((\textrm{faire}\circ_{}\textrm{exploser})\circ_{}(\textrm{une}\circ_{}\textrm{bombe}))))))\rule[-.2ex]{0pt}{.9em} \vdash n\rule[-.2ex]{0pt}{.9em}
                    }{\infer[\bo\textit{Lex}\bc]{n\rule[-.2ex]{0pt}{.9em}
                         }{\textrm{suspect}\rule[-.2ex]{0pt}{.9em}}
                         &
                         \infer[\bo/\textit{E}\bc]{\textrm{qui}\circ_{}(\textrm{a}\circ_{}(\textrm{tenté}\circ_{}(\textrm{de}\circ_{}((\textrm{faire}\circ_{}\textrm{exploser})\circ_{}(\textrm{une}\circ_{}\textrm{bombe})))))\rule[-.2ex]{0pt}{.9em} \vdash n \bs_{}n\rule[-.2ex]{0pt}{.9em}
                         }{\infer[\bo\textit{Lex}\bc]{(n \bs_{}n) /_{}(np \bs_{}s_{\textit{main}})\rule[-.2ex]{0pt}{.9em}
                              }{\textrm{qui}\rule[-.2ex]{0pt}{.9em}}
                              &
                              \infer[\bo/\textit{E}\bc]{\textrm{a}\circ_{}(\textrm{tenté}\circ_{}(\textrm{de}\circ_{}((\textrm{faire}\circ_{}\textrm{exploser})\circ_{}(\textrm{une}\circ_{}\textrm{bombe}))))\rule[-.2ex]{0pt}{.9em} \vdash np \bs_{}s_{\textit{main}}\rule[-.2ex]{0pt}{.9em}
                              }{\infer[\bo\textit{Lex}\bc]{(np \bs_{}s_{\textit{main}}) /_{}(np \bs_{}s_{\textit{ppart}})\rule[-.2ex]{0pt}{.9em}
                                   }{\textrm{a}\rule[-.2ex]{0pt}{.9em}}
                                   &
                                   \infer[\bo/\textit{E}\bc]{\textrm{tenté}\circ_{}(\textrm{de}\circ_{}((\textrm{faire}\circ_{}\textrm{exploser})\circ_{}(\textrm{une}\circ_{}\textrm{bombe})))\rule[-.2ex]{0pt}{.9em} \vdash np \bs_{}s_{\textit{ppart}}\rule[-.2ex]{0pt}{.9em}
                                   }{\infer[\bo\textit{Lex}\bc]{(np \bs_{}s_{\textit{ppart}}) /_{}(np \bs_{}s_{\textit{deinf}})\rule[-.2ex]{0pt}{.9em}
                                        }{\textrm{tenté}\rule[-.2ex]{0pt}{.9em}}
                                        &
                                        \infer[\bo/\textit{E}\bc]{\textrm{de}\circ_{}((\textrm{faire}\circ_{}\textrm{exploser})\circ_{}(\textrm{une}\circ_{}\textrm{bombe}))\rule[-.2ex]{0pt}{.9em} \vdash np \bs_{}s_{\textit{deinf}}\rule[-.2ex]{0pt}{.9em}
                                        }{\infer[\bo\textit{Lex}\bc]{(np \bs_{}s_{\textit{deinf}}) /_{}(np \bs_{}s_{\textit{inf}})\rule[-.2ex]{0pt}{.9em}
                                             }{\textrm{de}\rule[-.2ex]{0pt}{.9em}}
                                             &
                                             \infer[\bo/\textit{E}\bc]{(\textrm{faire}\circ_{}\textrm{exploser})\circ_{}(\textrm{une}\circ_{}\textrm{bombe})\rule[-.2ex]{0pt}{.9em} \vdash np \bs_{}s_{\textit{inf}}\rule[-.2ex]{0pt}{.9em}
                                             }{                                                  \infer[\bo/\textit{E}\bc]{\textrm{faire}\circ_{}\textrm{exploser}\rule[-.2ex]{0pt}{.9em} \vdash (np \bs_{}s_{\textit{inf}}) /_{}np\rule[-.2ex]{0pt}{.9em}
                                                  }{\infer[\bo\textit{Lex}\bc]{((np \bs_{}s_{\textit{inf}}) /_{}np) /_{}(np \bs_{}s_{\textit{inf}})\rule[-.2ex]{0pt}{.9em}
                                                       }{\textrm{faire}\rule[-.2ex]{0pt}{.9em}}
                                                       &
\infer[\bo\textit{Lex}\bc]{np \bs_{}s_{\textit{inf}}\rule[-.2ex]{0pt}{.9em}
                                                       }{\textrm{exploser}\rule[-.2ex]{0pt}{.9em}}
                                                  }
                                                  &
                                                  \infer[\bo/\textit{E}\bc]{\textrm{une}\circ_{}\textrm{bombe}\rule[-.2ex]{0pt}{.9em} \vdash np\rule[-.2ex]{0pt}{.9em}
                                                  }{\infer[\bo\textit{Lex}\bc]{np /_{}n\rule[-.2ex]{0pt}{.9em}
                                                       }{\textrm{une}\rule[-.2ex]{0pt}{.9em}}
                                                       &
\infer[\bo\textit{Lex}\bc]{n\rule[-.2ex]{0pt}{.9em}
                                                       }{\textrm{bombe}\rule[-.2ex]{0pt}{.9em}}
                                                  }
                                             }
                                        }
                                   }
                              }
                         }
                    }
               }
          }
     }
}
$
\end{adjustbox}
\hspace{0.15em}$\scriptstyle\rightarrow$\hspace{0.15em}
\begin{adjustbox}{max width=0.1785\linewidth,valign=c}
$
\left \langle\left \{\raisebox{2.4ex}{\,\setlength{\tabcolsep}{8pt}\begin{tabular}[t]{|l|} \firsthline
$y_{0}$ \\ \hline
$\textrm{police}(y_{0})$ \\[0pt] \lasthline  \end{tabular}\,}\right \} , \raisebox{2.4ex}{\,\setlength{\tabcolsep}{8pt}\begin{tabular}[t]{|l|} \firsthline
$d_{0}\ d_{1}\ e_{1}\ z_{0}$ \\ \hline
$\textrm{suspect}(z_{0})$ \\[0pt]
 $\textrm{tenter\_{}de}(d_{1} , z_{0} , e_{1})$ \\[0pt]
 $e_{1} : \raisebox{2.4ex}{\,\setlength{\tabcolsep}{8pt}\begin{tabular}[t]{|l|} \firsthline
$f_{1}\ e_{2}\ x_{2}$ \\ \hline
$\textrm{bombe}(x_{2})$ \\[0pt]
 $\textrm{faire}(f_{1} , z_{0} , x_{2} , e_{2})$ \\[0pt]
 $e_{2} : \raisebox{2.4ex}{\,\setlength{\tabcolsep}{8pt}\begin{tabular}[t]{|l|} \firsthline
$f_{2}\ x_{2}$ \\ \hline
$\textrm{exploser}(f_{2} , x_{2})$ \\[0pt] \lasthline  \end{tabular}\,}$ \\[10pt] \lasthline  \end{tabular}\,}$ \\[0pt]
 $\raisebox{2.4ex}{\,\setlength{\tabcolsep}{8pt}\begin{tabular}[t]{|l|} \firsthline
$d_{3}$ \\ \hline
$\textrm{temps}(d_{1})<\textrm{temps}(d_{3})$ \\[0pt]
 $\textrm{temps}(d_{3}) \circ \textrm{maintenant}$ \\[0pt] \lasthline  \end{tabular}\,}$ \\[0pt]
 $\textrm{relâcher}(d_{0} , y_{0} , z_{0})$ \\[0pt]
 $\raisebox{2.4ex}{\,\setlength{\tabcolsep}{8pt}\begin{tabular}[t]{|l|} \firsthline
$e_{3}$ \\ \hline
$\textrm{temps}(d_{0})<\textrm{temps}(e_{3})$ \\[0pt]
 $\textrm{temps}(e_{3}) \circ \textrm{maintenant}$ \\[0pt] \lasthline  \end{tabular}\,}$ \\[10pt] \lasthline  \end{tabular}\,}\right \rangle
 $
 \end{adjustbox}
\hspace{0.15em}$\scriptstyle\rightarrow$\hspace{0.15em}
    \begin{minipage}[c]{0.1885\linewidth}
    \begingroup
    \tiny
    \ttfamily
    \setlength{\baselineskip}{0.9\baselineskip}
exists a b c d e.(police(e) \& (suspect(c) \& (tenter\_de(b, c, a) \& (exists f g h.(bombe\_event(a, g) \& (faire\_event(a, h, c, g, f) \& exists i.(exploser\_event(f, i, g)))) \& (exists j.(<(temps(b), temps(j)) \& overlaps(temps(j), maintenant)) \& (relâcher(d, e, c) \& exists k.(<(temps(d), temps(k)) \& overlaps(temps(k), maintenant))))))))
    \par
    \endgroup
    \end{minipage}
\caption{Examples of lambda-term derivation, generated discourse and first-order representations for two French sentences. Each DRS separates
its presuppositional background from its main content; the second example also contains labelled
embedded content.}
\label{fig:examples_fol_generation_steps}
\end{figure}

Figure~\ref{fig:examples_fol_generation_steps} gives two successfully converted examples of sentences from the datasets, the steps of lambda-term derivation and DRS, and the resulting FOL formula.

\section{The FOLLIE resource}
\label{data:resource}

FOLLIE (for \underline{F}irst-\underline{O}rder \underline{L}ogic for \underline{L}anguage \underline{I}nference and \underline{E}ntailment for French)
associates French sentences with first-order representations while
preserving their premise--hypothesis organisation and inference labels.
It brings six complementary NLI collections into a common
language-to-logic resource. Its silver logical annotations are produced
by the construction pipeline, with a model-assisted completion layer for
selected FraCaS-FR and GQNLI-FR inputs. Sentence alignments support
semantic-parser development; their grouping into inference problems
supports reasoning over the resulting interpretations.
FOLLIE is publicly available under the CC BY-NC-SA 4.0 licence on Hugging Face\footnote{
\url{https://huggingface.co/datasets/maximoss/follie}.} and on GitHub.

\subsection{Natural-language sources and annotation provenance}
\label{data:provenance}

FraCaS \citep{Pulman1996d} contributes controlled semantic problems,
including multi-premise inferences with referential dependencies across
sentences. FOLLIE uses its French translation
\citep{amblard-etal-2020-french}. GQNLI
\citep{cui-etal-2022-generalized-quantifiers} targets generalised
quantifiers, while RTE3 \citep{giampiccolo-etal-2007-third} adds longer
premises from information extraction, retrieval, question answering and
summarisation, with three-way annotations from
\citet{de-marneffe-etal-2008-finding}. Their French versions,
GQNLI-FR and RTE3-FR \citep{skandalis-etal-2024-new}, were translated
manually.

SICK \citep{marelli-etal-2014-sick} provides short descriptions with
controlled lexical and syntactic variation. We translated it into French
with Helsinki-NLP's
\href{https://huggingface.co/Helsinki-NLP/opus-mt-tc-big-en-fr}{\texttt{opus-mt-tc-big-en-fr}}
model on 26 November 2023, then manually corrected the translations in
2025 using linguistic inspection and feedback from downstream parsing
and reasoning. XNLI \citep{conneau-etal-2018-xnli} contributes its
French evaluation sets with their broader range of genres. DACCORD
\citep{skandalis-etal-2023-daccord,skandalis-etal-2024-new} contributes
French sentences manually selected from AFP Factuel articles on the
Russian invasion of Ukraine, Covid-19 and climate change. Its manually
assigned and curated binary labels distinguish contradiction from
compatibility; compatibility includes both entailment and neutrality.

The sources combine authored French fact-checking text with translated
inference benchmarks, covering controlled and naturally occurring
descriptions. At the source-pair level, NLTK tokenisation
\citep{bird2009natural}, with punctuation retained, gives DACCORD mean premise and
hypothesis lengths of 34.53 and 26.72 tokens, compared with 21.50 and
10.80 for GQNLI-FR and 9.56 and 9.42 for the SICK-FR test set.
RTE3-FR has particularly asymmetric pairs:
development premises average 43.64 tokens and hypotheses 10.95; test
averages are 37.22 and 10.22. These source-span statistics characterise
the input collections before the pipeline's decomposition into sentence
units, separately from the aligned formula-slot population.

The lexical vocabulary of the representations retains French material,
including predicates such as \texttt{nommé}, \texttt{moitié} and
\texttt{plupart\_de}, alongside structural notation such as
\texttt{overlaps(temps(\ldots),maintenant)} and \texttt{subseteq}.
These representations combine French lexical composition with
first-order structure. Problem identities preserve the dependencies
between the interpretations of a premise set and its hypothesis.

Translation can preserve quantificational meaning while changing category:
GQNLI-FR renders determiner \emph{most} as nominal \emph{la plupart de},
while \emph{each}, \emph{every}, and \emph{neither} become \emph{chaque},
\emph{tout/tous les}, and \emph{aucun des deux}. These choices follow the
relevant French constructions \citep{amblard-etal-2020-french}.

\subsection{Instances, slots and retained coverage}
\label{data:coverage}

An aligned mapping is counted when a natural-language cell
(\texttt{p\emph{i}\_nl} or \texttt{h\emph{i}\_nl}) and its corresponding
representation cell (\texttt{p\emph{i}} or \texttt{h\emph{i}}) are both
non-empty after whitespace trimming. A mapping is an occurrence in a
particular problem: repeated sentences contribute multiple mappings,
and multi-premise problems can contribute more than two. Completeness
here measures alignment availability. Distinct sentence counts
deduplicate exact Unicode-normalised French text.

Table~\ref{data:composition} reports the retained row populations and
available mappings. FOLLIE contains fewer rows than some of its source
collections: DACCORD has 952 of the source's 1,034 pairs, each RTE3-FR
split has 790 of 800, and SICK-FR has 9,799 of 9,840. Formula
availability within these retained rows and end-to-end coverage of the
source collections therefore have different denominators.

\begin{table}[!tbp]
\centering
\small
\setlength{\tabcolsep}{3.5pt}
\resizebox{\linewidth}{!}{%
\begin{tabular}{llrrrrrr}
\hline
Source & Split & Rows & P mappings & H mappings & All mappings & NL types & P+H rows\\
\hline
DACCORD & test & 952 & 859 & 838 & 1,697 & 1,523 & 601\\
FraCaS-FR & test & 346 & 536 & 342 & 878 & 620 & 342\\
GQNLI-FR & test & 300 & 403 & 300 & 703 & 326 & 300\\
\multirow{2}{*}{RTE3-FR} & dev. & 790 & 814 & 775 & 1,589 & 1,399 & 592\\
 & test & 790 & 748 & 782 & 1,530 & 1,346 & 593\\
\multirow{3}{*}{SICK-FR} & train & 4,423 & 4,337 & 4,329 & 8,666 & 4,490 & 4,243\\
 & dev. & 492 & 484 & 483 & 967 & 876 & 475\\
 & test & 4,884 & 4,790 & 4,775 & 9,565 & 4,680 & 4,681\\
\multirow{2}{*}{XNLI-FR} & dev. & 2,312 & 1,671 & 2,091 & 3,762 & 2,646 & 1,383\\
 & test & 4,694 & 3,400 & 4,230 & 7,630 & 5,355 & 2,762\\
\hline
Total & & 19,983 & 18,042 & 18,945 & 36,987 & 18,819 & 15,972\\
\hline
\end{tabular}}
\caption{Composition of FOLLIE, including the additional FraCaS-FR and
GQNLI-FR completion layer. P+H rows contain at least one available premise
mapping and one available hypothesis mapping; they need not contain every
expected premise. NL types are deduplicated within each constituent file;
the total is deduplicated across all files and is therefore not their sum.}
\label{data:composition}
\end{table}

Across 19,983 NLI rows, FOLLIE contains 36,987 aligned mappings,
comprising 18,042 premise mappings and 18,945 hypothesis mappings.
At least one mapping is present on each side in 15,972 rows (79.928\%).
A further 1,098 retain premise mappings only,
2,899 retain hypothesis mappings only, and 14 retain neither. These last
rows are distributed between XNLI-FR (12) and DACCORD (two). Rows with
one-sided coverage still contribute sentence-to-representation instances.
Deductive NLI additionally requires checking every expected premise and
hypothesis: omitting a premise can change both entailment and
inconsistency.

Labels comprise 6,241 \texttt{yes}, 4,544 \texttt{no}, 9,187
\texttt{unknown} and 11 \texttt{undef} rows. Their interpretation
remains corpus-specific. In particular, DACCORD's compatible class
provides no separate gold judgement of neutrality, so inference
evaluation retains each source's task definition.

There are 18,819 distinct French text types and 18,871 distinct exact
text--representation pairs. Fifty-two texts occur with two distinct exact
representations. These cases include alternative analyses and deliberate
decomposition: a multi-clause source span can be repeated in two slots
whose representations express different clauses. A slot is consequently
an alignment unit whose interpretation requires its source span and
problem membership.

Repeated text is particularly consequential in SICK-FR. Pooling the
translated source's two sentence positions over all splits yields 19,680
occurrences but only 5,843 distinct exact French sentences. In FOLLIE,
74.98\% of the distinct SICK-FR test sentences also occur in training,
and 3,507 complete text--representation pairs are exact train--test
duplicates. This overlap follows the source benchmark's reuse of
sentences in different pairings. Its inherited splits can support
comparison with NLI work using the same protocol, but an NL-to-FOL
generalisation experiment requires a split that groups repeated sentence
types and related decompositions together. Otherwise, successful
prediction may reproduce an already observed mapping.

\subsection{Model-assisted completion}
\label{data:completion}

The completion layer supplies representations for 84 missing FraCaS-FR
slots and 27 missing GQNLI-FR slots, using GPT-5.5.
The FraCaS-FR additions comprise 50 premise slots and 34 hypothesis
slots; the GQNLI-FR additions comprise 27 hypothesis slots. Each request
supplied the corresponding dataset file, including the representations
already produced by the pipeline, and asked the model to propose
representations for the missing slots using the file's existing notation
and logical conventions. We used GitHub Copilot Chat
0.46.2 in Visual Studio Code 1.118.1, with the Xhigh thinking-effort
setting.

Before this completion, FraCaS-FR has 486 available premise mappings,
308 hypothesis mappings and 794 mappings overall. Completion raises
these counts to 536, 342 and 878, respectively, and raises the number of
rows with mappings on both sides from 292 to 342. For GQNLI-FR, the
premise count remains 403, the hypothesis count rises from 273 to 300,
and the overall count rises from 676 to 703; rows with mappings on both
sides rise from 273 to 300. The corresponding distinct-text counts rise
from 563 to 620 in FraCaS-FR and from 300 to 326 in GQNLI-FR.
Across FOLLIE, the base pipeline supplies
36,876 mappings and the completion layer supplies 111 more.

The additions supply candidate analyses for inputs that the construction
pipeline could not convert, including FraCaS ellipsis, group readings
and comparatives. Their model-generated provenance distinguishes them
from the analyses supported by categorial derivations.

\subsection{Distribution of representational structures}
\label{data:structures}

To characterise representation size, we count conjunction symbols
(\texttt{\&}) in each aligned mapping after masking complete single-quoted
tokens at lexical boundaries.\footnote{This lexical count does not require
a well-formed formula.} Table~\ref{data:profile} reports the resulting
distribution. Every statistic counts mapping occurrences, including
repeated source text. Means and standard deviations describe conjunction
counts; the quartiles reveal variation around the median, and
$\rho_s$ relates these counts to the corresponding text's
whitespace-token length.

\begin{table}[!tbp]
\centering
\small
\setlength{\tabcolsep}{3.5pt}
\resizebox{\linewidth}{!}{%
\begin{tabular}{llrrrrrrr}
\hline
Source & Split & Mappings & Total \texttt{\&} & Mean (SD) & Median [Q1,Q3] & Min. ($n$) & Max. ($n$) & $\rho_s$\\
\hline
DACCORD & test & 1,697 & 35,162 & 20.72 (10.24) & 19 [14,26] & 1 (1) & 97 (1) & .924\\
FraCaS-FR & test & 878 & 6,948 & 7.91 (5.03) & 7 [5,9] & 0 (1) & 56 (5) & .823\\
GQNLI-FR & test & 703 & 7,494 & 10.66 (5.28) & 10 [7,15] & 0 (3) & 27 (6) & .942\\
\multirow{2}{*}{RTE3-FR} & dev. & 1,589 & 28,519 & 17.95 (14.45) & 13 [8,25] & 0 (1) & 127 (1) & .966\\
 & test & 1,530 & 25,808 & 16.87 (13.18) & 11 [7,24] & 0 (3) & 89 (2) & .965\\
\multirow{2}{*}{SICK-FR} & train & 8,666 & 59,250 & 6.84 (3.66) & 6 [4,9] & 1 (56) & 27 (1) & .898\\
 & dev. & 967 & 6,830 & 7.06 (3.97) & 6 [4,9] & 1 (5) & 25 (1) & .901\\
 & test & 9,565 & 65,379 & 6.84 (3.63) & 6 [4,9] & 1 (78) & 27 (1) & .900\\
\multirow{2}{*}{XNLI-FR} & dev. & 3,762 & 52,220 & 13.88 (8.94) & 11 [7,18] & 0 (23) & 73 (3) & .913\\
 & test & 7,630 & 103,919 & 13.62 (9.13) & 11 [7,17] & 0 (29) & 149 (3) & .910\\
\hline
Total & & 36,987 & 391,529 & 10.59 (8.57) & 8 [5,13] & 0 (60) & 149 (3) & .906\\
\hline
\end{tabular}}
\caption{Conjunction-count distribution in the 36,987 aligned mappings.
SD is the population standard deviation; Q1 and Q3 are nearest-rank
quartiles. Parenthesised $n$ gives the number of occurrences attaining
each extreme. $\rho_s$ is Spearman's correlation with whitespace-token
length.}
\label{data:profile}
\end{table}

FOLLIE contains 391,529 conjunction symbols. The distribution is
right-skewed: its mean is 10.59 per mapping, its median is eight, and
its 95th percentile is 28. Only 60 mappings contain no conjunction.
DACCORD has the largest mean, whereas SICK-FR has smaller and similar
means across its three splits. Long-tail behaviour remains visible within
sources: RTE3-FR development has a maximum of 127 conjunctions despite
a median of 13. The overall maximum of 149 comes from one XNLI-FR
test premise repeated in three rows.

Conjunction counts correlate with whitespace-token length (split-level
Spearman $\rho_s=.823$--$.966$; pooled $.906$), as participant and modifier
conditions accumulate. They quantify representation size; a scope change
can alter entailment without adding many atoms.

\subsubsection{Connective, quantifier and relational incidence}
\label{data:incidence}

The other observed truth-functional notations are negation
(\texttt{not(...)}), disjunction (\texttt{|}) and implication
(\texttt{->}). Table~\ref{data:operators} counts aligned mappings
containing each notation at least once. Quantifiers and relational
notation are counted separately. The equality column includes
\texttt{neq(...)}; order-comparison incidence counts unquoted
\texttt{<} or \texttt{>} symbols, excluding the arrow in \texttt{->}. Bare question marks are
underspecification markers, whereas quoted values such as
\texttt{'context?'} do not contribute to that column.

\begin{table}[!tbp]
\centering
\small
\setlength{\tabcolsep}{3pt}
\resizebox{\linewidth}{!}{%
\begin{tabular}{lrrrrrrrrrr}
\hline
Source/split & Mappings & $\neg$ & $\lor$ & $\to$ & Any Boolean & $\exists$ & $\forall$ & Eq./neq & $<$/$>$ & Bare \texttt{?}\\
\hline
DACCORD & 1,697 & 379 & 61 & 140 & 442 & 1,697 & 118 & 1,155 & 1,445 & 200\\
FraCaS-FR & 878 & 21 & 13 & 89 & 108 & 877 & 73 & 330 & 639 & 41\\
GQNLI-FR & 703 & 66 & 18 & 6 & 87 & 703 & 3 & 550 & 386 & 33\\
RTE3-FR dev. & 1,589 & 58 & 60 & 27 & 127 & 1,589 & 25 & 932 & 1,221 & 144\\
RTE3-FR test & 1,530 & 58 & 41 & 28 & 107 & 1,530 & 25 & 834 & 1,198 & 133\\
SICK-FR train & 8,666 & 957 & 8 & 92 & 965 & 8,666 & 58 & 1,908 & 1,790 & 173\\
SICK-FR dev. & 967 & 98 & 0 & 10 & 98 & 967 & 5 & 194 & 196 & 22\\
SICK-FR test & 9,565 & 1,014 & 12 & 97 & 1,020 & 9,565 & 69 & 2,119 & 2,027 & 194\\
XNLI-FR dev. & 3,762 & 570 & 235 & 279 & 846 & 3,762 & 238 & 2,237 & 2,867 & 774\\
XNLI-FR test & 7,630 & 1,124 & 399 & 620 & 1,705 & 7,623 & 501 & 4,566 & 5,815 & 1,519\\
\hline
Total & 36,987 & 4,345 & 847 & 1,388 & 5,505 & 36,979 & 1,115 & 14,825 & 17,584 & 3,233\\
\hline
\end{tabular}}
\caption{Incidence of logical and relational notation in aligned
mappings. Each cell counts mappings containing the notation at least
once; columns are non-exclusive. ``Any Boolean'' is the union of
negation, disjunction and implication, excluding conjunction.}
\label{data:operators}
\end{table}

Negation, disjunction and implication occur in 4,345, 847 and 1,388 mappings,
respectively; 5,505 mappings (14.884\%) contain at least one of these
three. These incidence counts are non-exclusive. The corresponding token
counts are 4,636 negations, 1,159 disjunctions and 1,436 implications,
showing that some representations contain repeated operators.
Existential quantification appears in 36,979 mappings, and universal
quantification in 1,115 (3.015\%). Every representation containing an
explicit universal also contains an implication, consistent with the
restricted-universal pattern $\forall x(P(x)\to Q(x))$.
XNLI-FR contributes most disjunction and implication incidence: the
development and test files together account for 634 of the 847
disjunction-containing mappings and 899 of the 1,388
implication-containing mappings. SICK-FR, despite contributing 19,198
mappings, contains only 20 with disjunction; its non-conjunctive
structure predominantly involves negation. Relational notation is also
substantial: equality or \texttt{neq} occurs in 14,825 mappings and
order-comparison notation in 17,584. These features arise in the
representation of participants, plural groups, cardinalities and temporal
relations as well as in overt numerical expressions.

\subsubsection{Scope alternation}
\label{data:scope}

Scope analysis distinguishes nested operator dependencies from
operators occurring in independent branches. We parse formula strings
and put their Boolean
structure into negation normal form (NNF): implications become
disjunctions with negated antecedents, and negation is pushed through
connectives and quantifiers. No prenex conversion or Skolemisation is
performed, so local scope remains visible. For a root-to-leaf path
$\pi$ through the resulting formula tree, let $Q(\pi)$ be its sequence
of quantifier kinds, with entries in $\{\exists,\forall\}$, and
$B(\pi)$ its sequence of Boolean connectives, with entries in
$\{\land,\lor\}$. Define
\[
\begin{aligned}
a_Q(\varphi)&=\max_\pi\sum_{i=1}^{|Q(\pi)|-1}
  \mathbf 1[Q_i\ne Q_{i+1}],\\
a_B(\varphi)&=\max_\pi\sum_{i=1}^{|B(\pi)|-1}
  \mathbf 1[B_i\ne B_{i+1}],
\end{aligned}
\]
where empty sums are zero. For example,
$\forall x\exists y\,R(x,y)$ has $a_Q=1$, whereas
$(\forall xP(x))\land(\exists yQ(y))$ has $a_Q=0$: its
quantifiers occur in separate branches. Similarly,
$P\land(Q\lor R)$ has one Boolean switch on a relevant path.
NNF dualises quantifiers under negation, so alternation incidence can
differ from explicit \texttt{all} incidence in the input formulas.

Table~\ref{data:scope-profile} gives the scope profile of each corpus.
For each row, $N_s$ is the number of formula occurrences included in
this analysis. Writing $\varphi_j$ for its $j$th parsed formula
occurrence, the percentages are
$\frac{100}{N_s}\sum_{j=1}^{N_s}\mathbf 1[a_Q(\varphi_j)>0]$ and
$\frac{100}{N_s}\sum_{j=1}^{N_s}\mathbf 1[a_B(\varphi_j)>0]$, respectively.
Scope analysis counts formulas independently of NL-cell availability.
FraCaS-FR has four non-empty hypothesis formulas whose associated NL
cells are empty, giving 882 formula occurrences but 878 aligned
mappings. Conversely, three GQNLI-FR hypothesis strings have an extra
closing parenthesis, giving 703 aligned mappings and 700 parsed
formulas. The RTE3-FR development and XNLI-FR test profiles use
1,587 and 7,627 formulas, respectively, compared with 1,589 and 7,630
aligned mappings. These remaining populations retain their own $N_s$
denominators; their structural percentages do not supply parsing-success
rates for the aligned-mapping population.

\begin{table}[!tbp]
\centering
\small
\setlength{\tabcolsep}{5pt}

\begin{tabular}{lrrrrr}
\hline
Source/split & $N_s$ & $a_Q>0$ (\%) & $\max a_Q$ & $a_B>0$ (\%) & $\max a_B$\\
\hline
DACCORD & 1,697 & 20.27 & 3 & 23.22 & 4\\
FraCaS-FR & 882 & 9.41 & 2 & 11.68 & 4\\
GQNLI-FR & 700 & 7.71 & 1 & 10.00 & 1\\
RTE3-FR dev. & 1,587 & 3.84 & 2 & 7.37 & 4\\
RTE3-FR test & 1,530 & 3.92 & 2 & 6.21 & 3\\
SICK-FR, all splits & 19,198 & 9.59 & 1 & 7.09 & 3\\
XNLI-FR dev. & 3,762 & 15.87 & 4 & 20.23 & 4\\
XNLI-FR test & 7,627 & 15.84 & 3 & 20.09 & 4\\
\hline
\end{tabular}
\caption{Quantifier and Boolean scope alternation after NNF conversion.
$N_s$ counts the formulas included in the structural analysis, separately
from the complete aligned-mapping population. Each percentage uses the
$N_s$ in its own row. SICK-FR pools all three splits.}
\label{data:scope-profile}
\end{table}

The profile distinguishes structural variation from descriptive size.
DACCORD and XNLI-FR have deeper quantifier alternation, with maxima of
three and four switches, respectively; SICK-FR has no analysed path
with more than one. DACCORD also has the largest incidence of both
quantifier and Boolean alternation, while RTE3-FR has relatively rare
quantifier switching despite its comparatively large conjunction counts.
The two XNLI-FR splits have closely similar alternation incidence,
although their maximal quantifier depths differ. This provides a useful
description of the kinds of scope structures an inference system will
encounter, without identifying either alternation index with proof-search
complexity.

The profile also distinguishes quantifier vocabulary from the
quantificational structure available to a prover. GQNLI-FR contains only three
representations with an explicit universal quantifier, despite its focus
on generalised quantification. Some determiners survive through lexical
predicates and group representations, which require further semantic
axioms before their inferential properties become available to a prover.
The presence of a predicate naming a proportional quantifier does not
supply its intended truth conditions. Moreover, 3,233 mappings retain a
bare question-mark marker requiring resolution or explicit handling by
solver-facing normalisation.

The combined profiles locate complementary challenges: descriptive size
in long fact-checking and application-derived texts, scope dependencies
across diverse genres, and lexicalised quantifier meanings in controlled
tests. FOLLIE connects these linguistic properties to the
premise--hypothesis configurations on which automated inference operates.

\section{Automated reasoning over FOLLIE}
\label{reason:section}

We evaluate first-order inference over the FraCaS-FR, SICK-FR and GQNLI-FR
components of FOLLIE. The architecture combines preprocessing representation rewriting and
local background generation with automated inference \citep{rt_thesis}. Prover9 and Mace4
\citep{prover9-mace4} provide proof search and finite-model checks,
respectively, through NLTK \citep{bird-loper-2004-nltk}. Every committed decision has a Prover9
certificate; unsuccessful search yields an operational abstention.

\subsection{Working theories and inference obligations}
\label{reason:theory}

Let $P=\{P_1,\ldots,P_k\}$ be the formulas for an item's premises and $H$
the hypothesis formula. Eventualities are first-order individuals; verbal
predicates can take event arguments as in the construction pipeline.
$\mathsf{temps}(e)$ denotes an event's temporal trace and
$\mathsf{overlaps}(\mathsf{temps}(e),\mathsf{maintenant})$ anchors it to the
reference time. Predicate names, argument positions and parser markers
interface with semantic construction. Transformations act on in-memory copies,
preserving the released formulas.

Write $P^\star=\{P_1^\star,\ldots,P_k^\star\}$ and $H^\star$ for the processed inputs. Most processing
operations act independently on a formula; comparative separation also uses
the premise--hypothesis configuration. A generator collection supplies a
finite set of closed background axioms $\mathcal B(P,H)$, selected from the
local logical vocabulary and predicate--argument structures. The working
premise theory is
\begin{equation}
 \Theta(P,H)=P^\star\cup\mathcal B(P,H).
 \label{reason:theta}
\end{equation}
Generators inspect formula structure and lexical relations, without consulting
problem identifiers or gold labels. Execution requires no interactive proof
guidance or per-item human verdict; development did examine the benchmarks.

The classical inference obligations are familiar from discourse consistency
checking and logic-based textual entailment
\citep{blackburn:inria-00001208,bos-markert-2005-recognising,bos-2014-place}.
For a consistent working theory with independently licensed background assumptions,
entailment and contradiction correspond respectively to the
unsatisfiability of $\Theta\cup\{\neg H^\star\}$ and
$\Theta\cup\{H^\star\}$. The engine attempts three routes, in order:
\begin{align}
 E &: \quad \Theta(P,H)\vdash H^\star;\label{reason:E}\\
 C_1 &: \quad \Theta(P,H)\vdash\neg H^\star;\label{reason:C1}\\
 C_2 &: \quad \{H^\star\}\cup\mathcal B(P,H)\vdash F,
       \quad\text{for a premise }P_i^\star=\neg F.
       \label{reason:C2}
\end{align}
Here $\vdash$ records Prover9 closure within the applicable budget;
$E,C_1,C_2$ also denote the corresponding closure events. If $E$
does not close, $C_1$ targets $\neg H^\star$ directly, whereas $C_2$ proves
joint inconsistency through a consequence excluded by an explicitly negated
premise. The latter recognises only the exact outer serialisation
\texttt{-(...)}; equivalent negations need not trigger it. Reversing the
obligation can affect search through different placement of existential witnesses.

A candidate entailment returns \textsf{yes} if accepted and
\textsf{unknown} if its premise guard rejects it; neither case continues to
$C_1$ or $C_2$.

\subsection{Consistency guards and abstention}
\label{reason:guards}

An inconsistent premise theory entails both hypothesis polarities. For example,
encoding both colours of a brown-and-white dog becomes inconsistent if an
injected constraint incorrectly makes the colour predicates disjoint.

The \emph{premise guard} checks candidate entailments and contradictions for
positive evidence that the augmented premise theory is inconsistent. Its
first stage is a string-level trigger for recognised negative literals,
including negative consequents of generated disjointness axioms. A purely
positive relational theory without disequality, falsum, constraining equality
or function equations admits an all-true interpretation. However, the
implemented string test is only an approximation to that fragment: it does
not establish satisfiability for every untriggered theory. When triggered,
Mace4 first seeks a finite model of $\Theta$. A model cancels the proposed
downgrade. If no model is found within the bound, Prover9 attempts to derive
an explicit contradictory probe,
\begin{equation}
 \mathsf{zzc\_vac\_probe}(\mathsf{zzc\_a})\land
 \neg\mathsf{zzc\_vac\_probe}(\mathsf{zzc\_a}).
 \label{reason:probe}
\end{equation}
Only successful proof of this contradiction rejects the candidate. A Mace4
failure or a Prover9 timeout does not.

The \emph{hypothesis guard} addresses a related but different pathology.
Suppose translation fuses two participants and produces
$\exists x(\mathsf{enfant}(x)\land\mathsf{adulte}(x))$. Under a disjointness
axiom, the hypothesis is refutable independently of the premises. The guard
tests
\begin{equation}
 \{H^\star\}\cup\mathcal L^-\vdash\bot,
 \label{reason:hguard}
\end{equation}
where $\mathcal L^-$ contains the selected background axioms with negative
consequents. It applies only to candidate contradictions. A finite Mace4
model again cancels a downgrade; a positive Prover9 inconsistency proof
establishes it. Positive transfer axioms, positive premise-derived axioms and
arithmetic constraints with a falsum consequent are excluded from
$\mathcal L^-$. This choice protects premise-dependent arithmetic
contradictions, but makes the guard incomplete: it can miss an inconsistency
that requires a positive bridge followed by a negative constraint.

This downgrade is an evaluation policy: participant fusion or inappropriate
lexical opposition should not refute a satisfiable natural-language hypothesis.
Classical logic would legitimately refute a hypothesis that contradicts valid
lexical knowledge.

Let $V_P$ denote a positive inconsistency result from the premise guard and
$V_H$ one from the hypothesis guard. With $C=C_1\lor C_2$, the ordered
labelling function is
\begin{equation}
 \lambda(P,H)=
 \begin{cases}
 \textsf{unknown} & E\land V_P,\\
 \textsf{yes} & E\land\neg V_P,\\
 \textsf{unknown} & \neg E\land C\land(V_P\lor V_H),\\
 \textsf{no} & \neg E\land C\land\neg V_P\land\neg V_H,\\
 \textsf{unknown} & \neg E\land\neg C.
 \end{cases}
 \label{reason:lambda}
\end{equation}
The negations of $V_P$ and $V_H$ mean absence of an inconsistency proof,
including for untriggered or timed-out checks; they are not consistency
certificates.

Mace4 supplies satisfiability witnesses, never NLI labels. Bounded model-search
failure is inconclusive; only a Prover9 inconsistency certificate rejects a
candidate. Thus \textsf{unknown} combines unclosed target proofs and
guard-rejected proofs, without establishing semantic neutrality.

\subsection{Representation-sensitive preprocessing}
\label{reason:preprocessing}

Before inference, global signature normalisation separates overloaded symbols.
The parser can use a spelling such as \texttt{temps} both for a function and
for a predicate, or assign a relation several arities. Prover9 requires these
uses to be distinguished, a practical interface issue also encountered by
\citet{olausson-etal-2023-linc}. The normaliser renames non-baseline relational
arities with an arity suffix and prefixes a relation that shares a function
name with \texttt{pred\_}. Argument lists are preserved, and distinct
function/relation or arity uses receive distinct symbols. The signature
inventory is extracted by regular-expression matching over formula strings.

Many generated analyses have an existential prefix above conjoined atoms,
suitable for positive intersective descriptions. Operator scope requires
more: conjunction elimination can discard a negation, upper-bound, duration
or intensional marker like an optional modifier. The remedies comprise token
and binder hygiene, parser-conditioned reconstruction, and deliberate
abstraction of inadequately represented distinctions
(Table~\ref{reason:rewrites}).

\begin{table}[!tbp]
\centering
\small
\begin{tabular}{@{}>{\raggedright\arraybackslash}p{.23\linewidth}>{\raggedright\arraybackslash}p{.33\linewidth}>{\raggedright\arraybackslash}p{.36\linewidth}@{}}
\hline
Operation & Recognised configuration & Representation effect\\
\hline
Negative-determiner lift & \texttt{aucun}$(d)$ and \texttt{des}$(d,c)$ & Replace an asserted marker by negation of the designated existential scope\\
Upper-bound lift & \texttt{plus}, \texttt{a\_}, exact \texttt{num} marker & Exclude a relevant group above the cardinal threshold\\
Sentence-negation repair & Lifted sortals, one negative existential, witness linker & Restore the selected negative-existential clause reading\\
Restrictor annotation & Vague quantifier with associated sortals & Add an exact restrictor signature; original conjuncts remain\\
Comparative separation & Gradable head in a measure comparison & Distinguish comparative from independently asserted positive predication\\
Duration abstraction & Number, time unit and duration role & Replace the measure frame by a sealed duration atom\\
Universal repair & Stranded restrictor or existential implication & Reconstruct a restrictive universal under the parser convention\\
Signature and token hygiene & Arity overload, truth marker, duplicate binder & Separate symbols and interpret designated representation artefacts\\
\hline
\end{tabular}
\caption{Principal preprocessing operations, subject to the semantic
preconditions specified in the text.}
\label{reason:rewrites}
\end{table}

\paragraph{Negation and upper bounds.}
A recognised \texttt{aucun}/\texttt{des} construction removes those marker
atoms and negates the remaining existential body. This reconstruction is
licensed when that body is exactly the determiner's nuclear scope; negating
independent discourse content in the same conjunction would be too strong.
For \emph{au plus $N$}, the designated group-referent encoding is rewritten
schematically as
\begin{equation}
 \neg\exists\vec z\,
  (\mathsf{gt}(\mathsf{num}(c),N)\land R(c)\land V(\vec z)).
 \label{reason:upper}
\end{equation}
Here, $\vec z$ includes $c$. This expresses the selected upper-bound reading only if $c$ denotes a group
of the relevant participants, $\mathsf{num}$ represents its cardinality, and
$R,V$ preserve the original restrictor and event scope.

A separate sentence-negation repair targets participant sortals incorrectly
lifted above a negated existential. Its trigger requires a top-level
conjunction with exactly one negative existential, unary sortals on outer
variables as all remaining conjuncts, occurrence of those variables inside
the negation, absence of an internal cardinality or quantifier marker, and a
parser-specific \texttt{existe} witness linker. Together these conditions
select the intended $\neg\exists$ construction and avoid collapsing genuine
$\exists x\neg V(x)$ readings. This repair fires on 623 SICK-FR items in the
reported run and on none in FraCaS-FR or GQNLI-FR; activations do not quantify
its independent contribution to decisions.
Additional normalisations interpret positive \texttt{pas}, \texttt{pas\_de}
and \texttt{faux} markers as negation over their designated scope.

\paragraph{Quantifier restrictors.}
Vague or proportional quantifiers require a comparison base
\citep{Szabolcsi_2010}: on proportional or context-dependent readings,
many British delegates need not constitute many delegates. Preprocessing records $d$'s unary restrictors and class-naming
atoms in a fresh compound atom $Q_{R_1\cdots R_m}(d)$ for exact restrictor
matching. The original $Q(d)$ and restrictor conjuncts remain, however, so
\begin{equation}
 Q(d)\land R_1(d)\land R_2(d)\land\psi
 \models Q(d)\land R_2(d)\land\psi
 \label{reason:restrictor}
\end{equation}
remains valid. Signature enrichment therefore cannot block conjunct elimination.

A designated stranded restrictor
$\mathsf{nomme}(x,T)\land\forall x\,\varphi$ can be reattached as
$\forall x(\mathsf{nomme}(x,T)\rightarrow\varphi)$; this operation does not
fire in the three final evaluations. A stronger repair transforms a parser
artefact $\exists\vec v(A\rightarrow B)$ into
$\forall\vec x(A\rightarrow\exists\vec y B)$, where $\vec x$ comprises
antecedent variables and $\vec y$ consequent-only variables. Both are
semantic repairs under designated parser invariants.

\paragraph{Comparative, temporal and event structure.}
A pair-level comparative operation renames gradable-head atoms in the
property slot of \texttt{is\_at} when they accompany a measure comparison.
The same renaming is applied to a comparative hypothesis. Entity standards
are excluded, and the operation is suppressed when a non-comparative premise
independently asserts the adjective. This blocks an inference from
\emph{faster} to \emph{fast} based solely on an inert positive atom while
retaining an independently supplied positive baseline. This signature
operation leaves degree semantics unspecified. The standalone comparative helper
is a no-op in the reported configuration.

Past-marked type and event-binding atoms are similarly separated from
unmarked present predication where persistence is not entailed. Duration
processing replaces a matched number--unit--role frame by an atom such as
\texttt{en\_heure\_2}; \texttt{durant}/\texttt{pendant} additionally marks
the governed event predicate as atelic. This preserves selected bare-event
consequences while preventing deletion of the duration's internal constraints,
but sacrifices their compositional numerical structure. It does not supply
the arithmetic needed to refute incompatible duration descriptions.

For \texttt{ou}, event scope is distributed over two alternative fillers for
the same event role. By parser convention, the designated \texttt{true} token
denotes logical truth, licensing its removal. Repeated
quantifier variables are removed. The preprocessing core is shared by axiom
collection and proof submission, generally aligning generator and prover
signatures.

\subsection{Locally generated background knowledge}
\label{reason:axioms}

Background generation addresses the gap between an uninterpreted logical
signature and a linguistically interpreted representation. The emitted
formulas include logical or definitional principles, parser-interface
conventions, lexical hypotheses and benchmark-oriented quantitative policies.
All have equal force as prover premises. A certificate establishes a
consequence of the resulting theory; linguistic justification of its repairs
and axioms remains a separate requirement. The following schemas specify
the principal families and their licensing conditions.

\paragraph{Structural and sortal bridges.}
The structural backbone connects existential witnesses, event times and
reified properties through clauses including
\begin{align}
 &\forall x\exists w\,\mathsf{existe}(w,x),\label{reason:witness}\\
 &\forall cb\bigl(\mathsf{existe}(c,b)\rightarrow
       \mathsf{overlaps}(\mathsf{temps}(c),\mathsf{maintenant})\bigr),
       \label{reason:witness-time}\\
 &\forall exy\bigl((\mathsf{is\_at}(e,x,y)\land S(x))
                         \rightarrow S(y)\bigr).
       \label{reason:copula}
\end{align}
Each locally detected sortal predicate $S$ instantiates a separate first-order
clause. Analogous copular clauses transfer
names and relational modifiers. The backbone also supplies temporal-overlap
symmetry, selected interval/subset bridges, order transitivity and local ground
comparisons. Witness import and anchoring occur in the augmented theory for the certified
past-existential paraphrase of FraCaS-FR~1, from \emph{Un Italien est devenu le
plus grand ténor du monde} to \emph{Il y a eu un Italien qui est devenu le plus
grand ténor du monde}, with truth-token removal and the complete backbone.
Copular transfer requires the designated identity-copula reading of \texttt{is\_at}.

Curated French subsumption and synonym pairs yield
$\forall x(A(x)\rightarrow B(x))$ or
$\forall x(A(x)\leftrightarrow B(x))$ when both predicates occur at matching
arity. Morphological bridges reconcile attested number or gender variants of
predicate names. Event-to-type projections connect compatible reified and
unary sortal analyses under argument-position restrictions. Such bridges are
particularly useful where lemmatisation preserves an inflected form.
For example, the \texttt{etudiant}/\texttt{etudiante} bridge lets the feminine
instance in \emph{Marie est une étudiante} satisfy the universal in \emph{Tout
étudiant a utilisé son ordinateur}, supporting FraCaS-FR~117.

A shared-name rule supplies
\begin{equation}
 \forall xy\bigl((\mathsf{nomme}(x,n)\land\mathsf{nomme}(y,n))
               \rightarrow x=y\bigr)
 \label{reason:names}
\end{equation}
for each locally repeated capitalised name constant $n$. This is functionality of a name tag,
not the unique-name assumption: the latter distinguishes different names.
Equation~\ref{reason:names} is justified only for a document-local rigid name
with one intended bearer. Homonyms, names of classes such as \emph{Nobel},
and capitalised common nouns violate that condition. The detector's reliance
on capitalisation is consequently a source of possible over-identification.
Description similarity alone generates no identity axiom.

\paragraph{Lexical relations and predicate decomposition.}
French resources from Open Multilingual Wordnet
\citep{bond-foster-2013-linking}, queried through NLTK, and an offline
JeuxDeMots index \citep{lafourcade:lirmm-00200883} supply
synonym, hypernym and opposition relations for differing local predicates.
The JDM index used here contains approximately 1.5 million \texttt{isa}
edges, 116 thousand antonymy edges and 4.6 thousand synonymy edges. A shared
synset yields a biconditional; a hypernym path yields a directional
implication; selected antonymy or incompatibility relations yield negative
consequents. Lookup is sense-agnostic.
Curated blocklists suppress known misleading pairs and all edges targeting
\texttt{homme}; remaining relations still require a compatible sense, argument
structure and context, particularly for gradable or occupational oppositions.
An attested WordNet bridge is
\begin{equation}
 \forall exy\bigl(\mathsf{pourchasser}(e,x,y)
                   \leftrightarrow\mathsf{poursuivre}(e,x,y)\bigr).
 \label{reason:chase}
\end{equation}
For GQNLI-FR~234 it preserves event and participant arguments across
\emph{Plus de cinq gars ont pourchassé deux écolières} and \emph{Plusieurs gars
ont poursuivi deux écolières}; the complete theory closes the entailment.

Compound predicates such as \texttt{cracher\_sur} are bridged to their base
verb and prepositional component when their signatures fit a recognised
construction. A transparent literal use can support
$\mathsf{cracher\_sur}(e,x,y)\rightarrow\mathsf{cracher}(e,x)$, whereas a
figurative use need not. The generator can emit both entity-sharing and
event-sharing PP attachments. Since the spelling fixes neither, admitting
both is a coverage policy requiring additional attachment assumptions.
A reverse biconditional requires the convention that the
compound and the base-plus-PP frame are definitionally the same analysis.
Cross-arity bridges similarly reconstruct designated generic-object slots,
subject to role restrictions. Finally,
$\forall hi\,\mathsf{narration}(h,i)$ renders the discourse link inert under
a conjunctive-metadata convention.

\paragraph{Quantifier and cardinality policies.}
The marker hierarchy includes selected links from \texttt{tout} to
\texttt{plupart\_de}, to \texttt{beaucoup\_de}, and to positive cardinality,
plus existential import for \texttt{tout}. These are conditional policies
for already reified, non-empty groups sharing a finite restrictor. They are
not unrestricted generalised-quantifier laws: a universal over an empty
restrictor has no existential import, and \emph{most} entails \emph{many}
only with a compatible contextual threshold. Complex universal restrictors
are reconstructed as
\begin{equation}
 \forall v\bigl((S(v)\land\exists\vec r\,R(v,\vec r))
                   \rightarrow\exists\vec s\,V(v,\vec s)\bigr).
 \label{reason:universal}
\end{equation}
Here $S$ is the nominal sort, $R$ its relational restrictor and $V$ the event
matrix. The structural split uses the final restrictor-relational atom to identify
the restrictor/scope boundary. In FraCaS-FR~103, \emph{Tous les gestionnaires de
l'APCOM ont une voiture de fonction} yields a universal whose antecedent is
\begin{equation}
 R_{\mathrm{APCOM}}(d)\equiv\mathsf{gestionnaire}(d)\land
       \exists a\bigl(\mathsf{APCOM}(a)\land\mathsf{de}(d,a)\bigr).
 \label{reason:APCOM}
\end{equation}
Writing $V(d)$ for the existential car-having matrix, after name and copular
alignment the proof core is
$\forall d(R_{\mathrm{APCOM}}(d)\rightarrow V(d))$ together with the processed
instance $R_{\mathrm{APCOM}}(b)$ for the Durand witness supplied by \emph{Durand est un gestionnaire
de l'APCOM}: instantiation and modus ponens give $V(b)$, supporting the hypothesis
\emph{Durand a une voiture de fonction}. The available downward-restrictor
family does not trigger in the reported runs.

Numeric generators assert distinctness of different mentioned numeral
constants and the locally needed order facts. In our mathematical notation,
for mentioned $n_1<n_2$ they can supply
\begin{equation}
 \forall x\bigl(\mathsf{num}(x)=n_2
                  \rightarrow\mathsf{gt}(\mathsf{num}(x),n_1)\bigr).
 \label{reason:order}
\end{equation}
Numerals, $\mathsf{num}$ and comparison predicates are uninterpreted
first-order symbols whose arithmetic constraints come from the generated
clauses. Exact-count bridges derive lower bounds and
\emph{plus de}/\emph{moins de} descriptions while preserving the event
frame. On FraCaS-FR~15, the theory includes the instance $n_2=3,n_1=1$ and
certifies the entailment from \emph{Au moins trois ténors prendront part au concert} to
\emph{Il y a des ténors qui prendront part au concert}; the concert
participation frame is retained.

Ratio and proportion generators compute locally specified numerical
comparisons before emitting first-order bridges. A ratio such as
$86.33:100$ can support the larger category's outnumbering the smaller only
under a common population and location. Complement-majority inference needs
an exhaustive partition. Share-conflict generators associate functional share
tags with unequal established/claimed values: on GQNLI-FR~77, the theory
refutes a two-thirds poverty claim against an established $36\%$ share, since
$36/100<2/3$, under the exact-share convention. A separate policy can produce a
falsum-consequent constraint when an exact percentage has no integer
realisation in an exact finite scene, as with $60\%$ of six individuals.
These constraints presuppose an aligned reference set and event, an exact
finite population total, exhaustive categories and exact rational shares.
An existential count constrains its introduced group, leaving other
individuals of the same sort possible.

\paragraph{Scalar, group and relational constraints.}
Scalar opposition is tied to a shared sortal comparison base: a large mouse
can be a small animal. The sortal scalar schema is
\begin{equation}
 \forall q_1q_2b\;\neg\bigl(F_S(q_1,b)\land F_S(q_2,b)
                         \land A(q_1)\land A'(q_2)\bigr).
 \label{reason:scalar}
\end{equation}
Here $F_S(q,b)$ links a quality description to its bearer relative to sort
$S$, and $A,A'$ denote a locally selected opposite pair of predicates.
The frame must fix a common comparison class and evaluation occasion;
neither has a separate argument. The named scalar family requires
the same document-local name on both sides, the same sort $S$ and its
$S_{\mathrm{event}}$ quality frame, and opposite scalar predicates split
between premise and hypothesis. Its bundled constraint closes
FraCaS-FR~204, \emph{Mickey est un petit animal} versus \emph{Mickey est un
grand animal}. The generic schema in~\ref{reason:scalar} and the separate
\texttt{seul}-uniqueness family do not trigger in the evaluated runs.

Group-to-member transfer includes
\begin{equation}
 \forall gme\bigl((\mathsf{groupe}(g)\land\mathsf{de}(g,m)
                   \land\mathsf{jouer}(e,g))\rightarrow\mathsf{jouer}(e,m)\bigr).
 \label{reason:group}
\end{equation}
It closes SICK-FR~580, from \emph{Un groupe de garçons joue \ldots} to
\emph{\ldots\ des garçons jouent \ldots}. Transfer is restricted to selected readings of
\texttt{jouer}, \texttt{jouer\_dans}, \texttt{dans} and \texttt{courir}.
The admitted readings exclude collective predicates such as
\emph{se rassembler} or \emph{être nombreux}
\citep{Link+1983+302+323,deVries2017}. Relational converse clauses permute
arguments for locally present \emph{sur/sous}, \emph{devant/derrière} and
\emph{avant/après}; symmetry is supplied for selected binary relations.
For instance, $\forall xy(\mathsf{sur}(x,y)\leftrightarrow
\mathsf{sous}(y,x))$ preserves the converse argument order. These clauses
require a stable sense and reference frame; no unrestricted spatial
\emph{dans} transitivity is generated. Event-to-state
bridges, such as becoming a property, winning a contest or causing death,
similarly require aligned participants and temporal intervals.

\subsection{Resource bounds and the scope of a certificate}
\label{reason:resources}

The default Prover9 limit and Python join bound for direct entailment and
contradiction are both 15 seconds. A structural trigger for at least three
premises and two user-side universal quantifiers raises the entailment limit
to 35 seconds and its join bound to 40 seconds; a global subprocess timer
still kills the prover after 35 seconds. This is a formula-shape proxy for
multi-premise universal reasoning. Reverse contradiction keeps a 15-second
prover limit but its caller cuts off the attempt after 10 seconds. Within each
consistency guard, Mace4 searches domains up to size 20, with a ten-second
internal limit and a five-second caller bound; the subsequent Prover9 probe
has a five-second limit and a seven-second join bound.

The wrappers return successful closure as a Boolean. Search exhaustion,
timeout, parsing or prover failure, and caught exceptions can therefore all
reach the caller as absence of proof. The evaluated procedure is a partial
inference mechanism with operational abstention.

Evaluation measures gold-relative commitment precision, proof coverage and
three-way accuracy, complemented by analysis of representations and proof
dependencies.

\section{Experimental evaluation}
\label{eval:section}

We evaluate the inference engine on FOLLIE's FraCaS-FR
\citep{amblard-etal-2020-french}, GQNLI-FR
\citep{skandalis-etal-2024-new} and SICK-FR
\citep{skandalis-etal-2025-neurosymbolic} components, testing multi-premise
inference, quantitative reasoning and scene-description inference,
respectively. Conversion defects in the generated representations remain
part of the inference input.

\subsection{Evaluation populations and protocol}
\label{eval:population}

We score the populations in Table~\ref{eval:table-population}, excluding
eleven \textsf{undefined} FraCaS inputs and evaluating all GQNLI pairs.
The SICK inference input contains 9,799 pairs; 400 lack a required premise
or hypothesis formula, leaving 9,399 evaluated pairs pooled across splits.
This population differs from the 9,840-pair French textual resource. We
also score 4,681 available SICK test pairs separately. The reasoning experiments and the 111 LLM-assisted completions have no
linked item-level version manifest; these results therefore do not isolate
the effect of completion on reasoning.

\begin{table}[!tbp]
\centering
\begin{adjustbox}{max width=\textwidth}
\small
\begin{tabular}{lrrrrr}
\hline
Corpus & Evaluated & Entailment & Contradiction & Neutral & Excluded\\
\hline
FraCaS-FR & 335 & 204 & 33 & 98 & 11\\
GQNLI-FR & 300 & 97 & 103 & 100 & 0\\
SICK-FR & 9,399 & 2,702 & 1,373 & 5,324 & 400\\
\hline
\end{tabular}
\end{adjustbox}
\caption{Inference evaluation populations and gold-label counts. FraCaS
exclusions have undefined labels; SICK exclusions lack a required formula.
SICK counts refer to the 9,799-pair inference input, not the entire textual
resource.}
\label{eval:table-population}
\end{table}

We apply the same proof-only decision policy to all three corpora.
Generators use neither problem identifiers nor gold labels; labels determine
scoring and undefined-item exclusion. We developed the lexicon, rewrites,
axiom families and safeguards through benchmark inspection. The evaluation
therefore measures an automatic inference policy under benchmark-informed
development.

\subsection{Metrics for proof-based classification}
\label{eval:metrics}

Let $\mathcal E$ be the evaluated set, $n=|\mathcal E|$, $g_i$ its gold
label and $\hat y_i$ the engine output. We encode entailment, contradiction
and neutrality as \textsf{yes}, \textsf{no} and \textsf{unknown}; the last
label also serves as the engine's abstention output. Define the committed set as
$\mathcal C=\{i\in\mathcal E:\hat y_i\in\{\textsf{yes},\textsf{no}\}\}$.
The three-way accuracy, commitment coverage and gold-relative commitment
precision are
\begin{align}
 \mathrm{Acc}&=\frac{1}{n}\sum_{i\in\mathcal E}\mathbf1[\hat y_i=g_i],
 &\rho&=\frac{|\mathcal C|}{n},
 &\pi&=\frac{\sum_{i\in\mathcal C}\mathbf1[\hat y_i=g_i]}{|\mathcal C|}.
 \label{eval:metrics-equation}
\end{align}
We also report macro-$F_1$ and support-weighted $F_1$, obtained by averaging
per-class $F_1$ uniformly and by gold support, respectively. Per-polarity
recall measures how many gold entailments or contradictions are proved;
per-polarity precision measures agreement among the corresponding
commitments.

These quantities separate two kinds of failure. A gold-discordant commitment
is a proved \textsf{yes} or \textsf{no} that differs from $g_i$. A recall
miss is \textsf{unknown} on a gold entailment or contradiction. Let $\Phi$
count discordant commitments and $\Phi_u$ count commitments on gold-neutral
items. Then
\begin{equation}
 \mathrm{Acc}=\frac{|\mathcal C|-\Phi+s_u-\Phi_u}{n},
 \label{eval:decomposition}
\end{equation}
where $s_u$ is neutral gold support. The first term rewards correct
commitments and the second correct uncommitted outputs. Accuracy alone
therefore obscures the different contributions of proof coverage and class
distribution. Commitment precision, in turn, measures agreement with an
annotation policy; it does not estimate the probability that every formula or
injected axiom is a faithful interpretation. 

\subsection{Aggregate performance}
\label{eval:aggregate}

We obtain 130 gold-correct decisions among 139 FraCaS commitments, 111
among 111 GQNLI commitments, and 1,396 among 1,415 SICK commitments
(Table~\ref{eval:table-aggregate}). This high agreement accompanies
substantially different coverage across the corpora.

\begin{table}[!tbp]
\centering
\begin{adjustbox}{max width=\textwidth}
\small
\setlength{\tabcolsep}{4pt}
\begin{tabular}{lrrrrrrr}
\hline
Corpus & Acc. & Macro-$F_1$ & Weighted $F_1$ & $\rho$ & $\pi$ & $R_{\mathrm{yes}}$ & $R_{\mathrm{no}}$\\
\hline
FraCaS-FR & 65.37 & 57.47 & 65.91 & 41.49 & 93.53 & 59.80 & 24.24\\
GQNLI-FR & 70.33 & 70.64 & 70.65 & 37.00 & 100.00 & 54.64 & 56.31\\
SICK-FR & 71.35 & 54.36 & 66.24 & 15.05 & 98.66 & 45.78 & 11.58\\
\hline
\end{tabular}
\end{adjustbox}
\caption{Final aggregate results, in percent. $\rho$ is commitment coverage
and $\pi$ agreement with gold among commitments; $R_{\mathrm{yes}}$ and
$R_{\mathrm{no}}$ are entailment and contradiction recall. Values are
computed from Table~\ref{eval:table-confusion}.}
\label{eval:table-aggregate}
\end{table}

The confusion matrices in Table~\ref{eval:table-confusion} identify where the
remaining errors arise. There are no entailment--contradiction reversals on
FraCaS or GQNLI; SICK has five. FraCaS's nine discordant commitments are
entailments on gold-neutral problems. SICK has fourteen commitments on
gold-neutral pairs in addition to its five reversals. Most errors are
abstentions on gold entailments or contradictions. Entailment and
contradiction precision are, respectively,
93.13\% and 100\% on FraCaS, 100\% and 100\% on GQNLI, and 99.44\% and
92.98\% on SICK. Contradiction recall is nevertheless only 11.58\% on
SICK and 24.24\% on FraCaS; it reaches 56.31\% on GQNLI, whose numeric
constraints support explicit incompatibility proofs.

\begin{table}[!tbp]
\centering
\begin{adjustbox}{max width=\textwidth}
\small
\begin{tabular}{llrrr}
\hline
Corpus & Gold label & Predicted yes & Predicted no & Predicted unknown\\
\hline
FraCaS-FR & yes & 122 & 0 & 82\\
 & no & 0 & 8 & 25\\
 & unknown & 9 & 0 & 89\\
\hline
GQNLI-FR & yes & 53 & 0 & 44\\
 & no & 0 & 58 & 45\\
 & unknown & 0 & 0 & 100\\
\hline
SICK-FR & yes & 1,237 & 3 & 1,462\\
 & no & 2 & 159 & 1,212\\
 & unknown & 5 & 9 & 5,310\\
\hline
\end{tabular}
\end{adjustbox}
\caption{Final confusion matrices. Gold-neutral agreement and operational
abstention share the output label \textsf{unknown}.}
\label{eval:table-confusion}
\end{table}

SICK attains the highest accuracy despite the lowest commitment coverage
because 5,310 of its 5,324 neutral items remain unknown. These correct
outputs contribute 56.50 percentage points to the 71.35\% accuracy. The
system also abstains on 1,462 gold entailments and 1,212 gold contradictions.
The resulting 54.36\% macro-$F_1$ shows why the accuracy cannot be read as
balanced inferential competence. FraCaS has the opposite distribution:
61\% of its evaluated items are entailments, so abstention has a larger
accuracy cost. GQNLI's near-balanced distribution makes its 70.33\%
accuracy less dependent on a dominant class.

\subsection{Semantic coverage and axiom use}
\label{eval:phenomena}

FraCaS's nine sections localise semantic strengths and weaknesses
(Table~\ref{eval:table-fracas}). Generalised quantifiers attain 81.1\%
accuracy; only the eight-item verb section scores higher. Temporal reference
has the lowest accuracy and commitment precision, accounting for four of
the nine discordant commitments. The aspectual losses examined in
Section~\ref{eval:fracas-errors} help explain this pattern. Five sections
have no discordant commitments: nominal anaphora, ellipsis, comparatives,
verbs and attitudes. Their coverage ranges from 35.5\% to 62.5\%, so the
unproved cases remain substantial even where every commitment agrees with
the gold labels.

\begin{table}[!tbp]
\centering
\begin{adjustbox}{max width=\textwidth}
\small
\begin{tabular}{lrrrrr}
\hline
FraCaS section & $n$ & Accuracy & Coverage & Precision & Discordant\\
\hline
Generalised quantifiers & 74 & 81.1 & 45.9 & 91.2 & 3\\
Plurals & 33 & 57.6 & 39.4 & 92.3 & 1\\
Nominal anaphora & 28 & 67.9 & 53.6 & 100.0 & 0\\
Ellipsis & 55 & 63.6 & 43.6 & 100.0 & 0\\
Adjectives & 22 & 68.2 & 45.5 & 90.0 & 1\\
Comparatives & 31 & 64.5 & 35.5 & 100.0 & 0\\
Temporal reference & 71 & 49.3 & 31.0 & 81.8 & 4\\
Verbs & 8 & 87.5 & 62.5 & 100.0 & 0\\
Attitudes & 13 & 69.2 & 38.5 & 100.0 & 0\\
\hline
\end{tabular}
\end{adjustbox}
\caption{FraCaS results by semantic section. Rates are percentages; the last
column counts discordant commitments.}
\label{eval:table-fracas}
\end{table}

Axiom generation varies with the semantic demands of the corpus
(Table~\ref{eval:table-profiles}). GQNLI receives 9.91 numeric and 29.62
structural axioms per evaluated item on average; the structural count includes
order facts needed by its cardinality and proportion vocabulary. FraCaS uses
more shared-name coreference and universalisation rules. SICK receives fewer
structural axioms but more lexical relations: curated hypernymy, WordNet and
JeuxDeMots together average 0.77 axioms per item, compared with 0.10 on
FraCaS and 0.40 on GQNLI. Compound decomposition is also most frequent on
SICK. These are injection profiles: proof-dependency analysis would be
needed to establish which generated axioms occur in successful proofs.

\begin{table}[!tbp]
\centering
\begin{adjustbox}{max width=\textwidth}
\small
\begin{tabular}{lrrr}
\hline
Axiom family & FraCaS-FR & GQNLI-FR & SICK-FR\\
\hline
Structural backbone & 10.02 & 29.62 & 3.41\\
Numeric distinctness, order and lifting & 0.31 & 9.91 & 0.06\\
Lexical/event interface & 1.53 & 1.40 & 0.21\\
Shared-name coreference & 0.70 & 0.09 & $<0.01$\\
Compound decomposition & 0.35 & 0.44 & 0.54\\
Curated hypernymy & 0.02 & 0.18 & 0.29\\
WordNet & 0.03 & 0.06 & 0.22\\
JeuxDeMots & 0.05 & 0.16 & 0.26\\
Structural temporal rules & 0.22 & 0.34 & 0.02\\
\emph{Tout}-universalisation & 0.16 & 0.00 & 0.00\\
Ratio, proportion and share & 0.00 & 0.15 & 0.00\\
\hline
\end{tabular}
\end{adjustbox}
\caption{Mean injected axioms per evaluated item for selected families.
Injection counts measure generator activation, not proof dependence or
isolated performance gain.}
\label{eval:table-profiles}
\end{table}

The accepted proof routes show a corresponding asymmetry: FraCaS yields
131 entailments and eight contradictions, GQNLI 53 and 58, and SICK 1,244
and 171. GQNLI supports a more balanced yield of the two proof polarities,
including numerical and proportional refutations. The first-order-heavy
budget extension activates on
29 FraCaS items and on no GQNLI or SICK item. The sentence-negation scope
repair activates on 623 SICK pairs and on neither controlled suite. Its
formula-based trigger is thus concentrated on SICK's conversion pattern;
activation counts alone leave its contribution to correct predictions
undetermined.

\subsection{Development checkpoints for consistency safeguards}
\label{eval:checkpoints}

We downgrade 40 SICK candidates through the premise-vacuity guard and ten
through the hypothesis-vacuity guard; neither guard downgrades a FraCaS or
GQNLI item. Three
successive checkpoints show fewer discordant commitments and increased
accuracy and precision (Table~\ref{eval:table-checkpoints}). The last
transition combines the hypothesis guard and lexical-edge blocklist,
precluding separate effect estimates. Without coverage at each checkpoint,
these developmental comparisons cannot establish that correct commitments
are preserved.

\begin{table}[!tbp]
\centering
\begin{adjustbox}{max width=\textwidth}
\small
\begin{tabular}{lrrr}
\hline
SICK configuration & Discordant & Precision & Accuracy\\
\hline
Negation repair, no vacuity guard & 62 & 95.78 & 71.08\\
Add premise-vacuity guard & 32 & 97.76 & 71.26\\
Add hypothesis guard and blocklist & 19 & 98.66 & 71.35\\
\hline
\end{tabular}
\end{adjustbox}
\caption{Successive SICK development checkpoints. Discordant counts
gold-discordant commitments; precision is agreement among commitments.
Rates are percentages.}
\label{eval:table-checkpoints}
\end{table}

SICK item 10 illustrates the cost of this policy. It describes a brown
dog attacking another animal in front of a tall man in trousers, while the
hypothesis drops \emph{tall}. Prover9 initially derives the weaker hypothesis.
The generated theory also contains unresolved \texttt{atomic\_sub} and
\texttt{unknown\_} structures, animal/man incompatibility and an
\texttt{empty\_intersect}-to-disequality convention. The premise guard
finds a positive Prover9 inconsistency proof after bounded model search is
unsuccessful and changes the candidate entailment to \textsf{unknown}.
Although the gold label is entailment, retaining an entailment proof from an
inconsistent theory would provide no discriminating evidence: the same theory
proves the opposite polarity. The downgrade loses one gold-correct
commitment while preserving the stated evidential policy.

\section{Error analysis and interpretation}
\label{eval:errors}

Residual failures involve conversion defects that change
scope, participant identity or argument structure. Representation gaps omit a
discourse, aspectual or quantitative distinction. Missing lexical knowledge
prevents proofs; excessive axioms can create wrong commitments. Search limits
cause abstention, and labels can conflict with defensible readings. These
causes interact, so mutually exclusive causal counts require controlled
intervention. The confusion matrices count outcomes; the analyses below
identify recurrent mechanisms.

\subsection{FraCaS: scope, restrictors and temporal interpretation}
\label{eval:fracas-errors}

The nine FraCaS discordant commitments all predict entailment for a
gold-neutral item. Item 22 pairs \emph{Aucun délégué n'a terminé le rapport à
temps} with the corresponding hypothesis lacking \emph{à temps}. The natural-language
premise permits a delegate to finish late. Schematically, with the report
argument suppressed, its intended structure is
\begin{equation}
 \neg\exists x,e\bigl(\mathrm{delegate}(x)\land
 \mathrm{finish}(e,x)\land\mathrm{onTime}(e)\bigr),
 \label{eval:negation-example}
\end{equation}
which does not entail
$\neg\exists x,e(\mathrm{delegate}(x)\land\mathrm{finish}(e,x))$.
In the generated representation the adjunct is placed outside the negation;
the stronger negative claim then follows by conjunct elimination. The proof
is valid for the formula, but the formula has lost the source contrast.

Further errors disconnect quantificational variables from their restrictors
or flatten a coordinated noun phrase. Generated structural transfer can then
make a hypothesis derivable whose natural-language quantifier dependency is
absent. Restrictor-signature enrichment does not by itself cure this defect:
adding a compound quantifier/restrictor atom while retaining a bare
quantifier marker and its conjuncts leaves conjunct elimination available.
The retained markers allow structural transfer to bypass the enriched
restrictor.

Temporal-reference errors retain an ordering between event times but lose
the aspectual conditions under which a before/after converse is licensed.
Accomplishments, achievements, activities, states and progressives have
different inferential behaviour in the FraCaS specification
\citep{Pulman1996d}. A single event-order relation cannot preserve all of
those distinctions. In adjective item 201, a partial past-time repair of
\emph{ancien} leaves a present \texttt{etudiant} predicate elsewhere in
the formula, preserving the erroneous consequence. Richer composition of
aspect and temporally restricted predication is needed to keep these
readings distinct.

Among abstentions, unresolved anaphoric markers such as
\texttt{masculin?} and \texttt{(a=?)} leave the hypothesis without a
licensed antecedent. Plural and elliptical analyses also retain
\texttt{atomic\_sub} placeholders or collapse different source problems
into the same logical form. When opposing relations receive the same
premise--hypothesis representation, no reasoner restricted to that input can
recover the erased distinction. Comparative misses reveal another gap:
fresh measure terms for two occurrences of the same comparison standard
block transitive comparative proofs until their identity is established.
Cross-interval counting, such as combining past and future visits, additionally
requires explicit interval disjointness and cardinality addition.

\subsection{GQNLI: systematic limits of quantitative representation}
\label{eval:gqnli-errors}

All 89 GQNLI errors are abstentions on gold entailments or contradictions.
Their distribution follows the dataset's repeated construction: each of
100 patterns occurs three times under lexical variation. Grouping problem
IDs by residue modulo 100, we find that 27 of the 34 error-affected groups
contain three errors; the remaining seven account for eight errors. This
concentration supports a
phenomenon-level diagnosis while leaving room for lexical and conversion
differences within a pattern.

Closed-scene counting is a major obstacle. An enumeration of six dogs and
their activities does not imply, under unrestricted open-world semantics,
that no other dogs exist. Summing coloured or activity-defined subgroups
also requires their disjointness. A statement that more than four dogs are
sitting, or exactly half are sitting, may follow under an exhaustive scene
convention without following from the available FOL. The closed-world module
is disabled in the evaluated configuration. Explicit completeness and
disjointness conditions could make such inferences available; treating every
enumeration as exhaustive would add information unsupported by the literal
theory.

Proportions present a different problem. A count over one population does
not determine a fraction over another unless their reference sets are
identified. The African-poverty patterns, for example, require connecting
the counted individuals to the total population denoted in the hypothesis.
An existential witness can denote a subset of that population, so its
cardinality cannot silently become the denominator of the population share.
We rejected a tested percentage-witness bridge because it licensed
conclusions over an insufficiently identified reference set. The required
repair is to represent the common population, subset relation and exact
rational quantity together.

Other unresolved patterns are concrete conversion losses: an interval from
one to two is rendered as a conjunction of two exact cardinalities; a mutual
relation reuses a participant variable; and a ratio hypothesis quantifies
over a population not identified by the premise's existential individuals.
These cases show why successful numeric axiom injection is conditional on
the preservation of the quantities and sets to which the axioms apply.

\subsection{SICK: scene identity and axiom interaction}
\label{eval:sick-errors}

SICK contains 2,693 incorrect evaluated outputs: 1,462 entailment abstentions,
1,212 contradiction abstentions and 19 discordant commitments. Scene
identity is a recurrent obstacle in both recall categories. Two descriptions
can refer to the same woman, instrument or event while their formulas
introduce separate existential witnesses. A positive event and a negative
event are not necessarily contradictory if their participants denote
different individuals. A lexical relation such as
\emph{trancher}$\rightarrow$\emph{couper} cannot supply that missing
identity. Entity and event coreference must be licensed by a discourse or
scene condition. Conversely, an unnecessary argument or fused participant can
block even a lexical entailment that otherwise has the required relation.

The nineteen discordant commitments reveal the risks of enriching imperfect
formulas. In item 7586, a JeuxDeMots-derived opposition between
\emph{cycliste} and \emph{motard} yields a wrong contradiction: the roles
are not mutually
exclusive properties of a person. Several other contradiction proofs combine
a shared copular/property variable, positive \texttt{is\_at} transfer and
selected gender-disjointness axioms. Reuse of a global \texttt{unknown\_}
constant forces unrelated participants to coincide in item 8931. Item 9369
treats the common noun \emph{État} as a proper name, enabling a
shared-name identity followed by a geographic incompatibility. In each
case, the certificate locates an actual consequence of the submitted theory;
the error concerns why that theory was supplied.

The restricted hypothesis guard cannot detect all of these interactions.
It tests negative-consequent lexical axioms after a candidate contradiction;
positive transfer axioms are omitted. It therefore misses inconsistencies
requiring a positive bridge before a disjointness rule. It also does not
inspect a candidate entailment. Item 2628 combines a fused
duration/participant variable with a lexical incompatibility, making the
existential content inside a negated hypothesis impossible. Extending the
checks to the complete relevant axiom set and to both asserted polarities is
a concrete prospective repair, but its precision and coverage effects have
not been evaluated here.

Some disagreements require joint review of annotation and translation.
Three neutral pairs receive identical premise
and hypothesis formulas; this establishes a conflict between representation
and label but cannot, by itself, distinguish a conversion collision from a
gold error. \emph{Cupcake}$\rightarrow$\emph{gâteau} is a plausible
taxonomic inclusion, whereas a bidirectional \emph{plage}/\emph{rivage}
bridge depends on sense and context. These cases call for independently
adjudicated readings. We retain the inherited labels in all scores.

\subsection{Comparison with other inference routes}
\label{eval:comparison}

We compare systems by language, evaluation population, logical-form
provenance and proof automation
(Tables~\ref{eval:table-logic-fracas}--\ref{eval:table-encoders}). Coq-based
inference includes automatic tactic search
\citep{mineshima-etal-2015-higher} and human-assisted proof construction over curated
semantic analyses \citep{chatzikyriakidis-bernardy-2019-wide,bernardy-chatzikyriakidis-2021-applied}.
Cross-language results additionally vary translation and lexical resources.
These comparisons provide descriptive context; assessing the French
inference routes directly requires paired decisions on identical rows.

\begin{table}[!tbp]
\centering
\begin{adjustbox}{max width=\textwidth}
\small
\setlength{\tabcolsep}{4pt}
\begin{tabular}{@{}>{\raggedright\arraybackslash}p{.25\linewidth}>{\raggedright\arraybackslash}p{.20\linewidth}>{\raggedright\arraybackslash}p{.15\linewidth}>{\raggedright\arraybackslash}p{.32\linewidth}@{}}
\hline
System & Population & Accuracy & Representation and proof construction\\
\hline
\citet{mineshima-etal-2015-higher}
 & EN: 181, six sections & 69 & Automatic CCG--HOL; automatic Coq tactics\\
FraCoq 2019\newline\citep{chatzikyriakidis-bernardy-2019-wide}
 & EN: 259, eight sections & 89 & Curated GF trees; dynamic semantics; problem-specific Coq proofs\\
FraCoq 2021\newline\citep{bernardy-chatzikyriakidis-2021-applied}
 & EN: 329 scored cases & 81 & Curated GF trees; interval semantics and arithmetic; human-assisted Coq proofs\\
\citet{Haruta_Mineshima_Bekki_2022}
 & EN: 73/22/31/13 & 99/95/90/92 & Automatic CCG--typed FOL; Vampire; abduction and parser-repair rules\\
FOLLIE + Prover9
 & FR: all 335, nine sections & 65.37 & Generated TLG--DRS--FOL; automatic proof-only decisions\\
\hline
\end{tabular}
\end{adjustbox}
\caption{FraCaS proof-based results (accuracy in percent). For
\citet{Haruta_Mineshima_Bekki_2022}, populations and scores are ordered as quantifiers, adjectives, comparatives
and attitudes; they concern the configuration with abduction and hand-coded
POS/lemma and multiword repairs.}
\label{eval:table-logic-fracas}
\end{table}

\citet{mineshima-etal-2015-higher} omit anaphora, ellipsis and temporal
reference. FraCoq \citep{bernardy-chatzikyriakidis-2021-applied} supplies
dynamic discourse semantics and reaches 73\% on temporal reference.
\citet{Haruta_Mineshima_Bekki_2022} supply degree variables, comparison
classes and arithmetic constraints before proof search. Their analysis
provides a concrete account of comparison standards that remain unresolved
in our French formulas (Section~\ref{eval:fracas-errors}), motivating richer
degree and reference-argument structure at the semantic interface.

\begin{table}[!tbp]
\centering
\begin{adjustbox}{max width=\textwidth}
\small
\setlength{\tabcolsep}{4pt}
\begin{tabular}{@{}>{\raggedright\arraybackslash}p{.46\linewidth}>{\raggedright\arraybackslash}p{.23\linewidth}rr@{}}
\hline
System & Evaluation & Accuracy & Precision\\
\hline
\citet{martinez-gomez-etal-2017-demand}
 & EN test: 4,927 & 82.97 & 97.07\\
\quad Add VerbOcean & EN test: 4,927 & 83.13 & 97.04\\
\citet{Haruta_Mineshima_Bekki_2022}
 & EN test: 4,927 & 82 & --\\
LangPro + abduction\newline\citep{abzianidze-2020-learning}
 & EN test: 4,927 & 84.4 & 94.3\\
LangPro + abduction\newline\citep{Abzianidze_Kogkalidis_2021}
 & NL test: 4,927 & 78.83 & --\\
French LangPro + abduction\newline\citep{skandalis-etal-2025-neurosymbolic}
 & FR test & 71.1 & 96.8\\
FOLLIE + Prover9 & FR available test: 4,681 & 71.31 & 98.57\\
\hline
\end{tabular}
\end{adjustbox}
\caption{SICK test results (percent). Precision retains each publication's
entailment/contradiction commitment denominator; dashes denote values not
reproduced here. LangPro's English and Dutch rows aggregate several parser
configurations. French LangPro uses abductively learned relations without a
dedicated French knowledge base.}
\label{eval:table-logic-sick}
\end{table}

\citet{martinez-gomez-etal-2017-demand} use failed proof subgoals to restrict lexical injection to
predicates sharing an argument; their base system has 62.13\% recall of
entailment/contradiction relations. This couples lexical selection with proof
obligations.
LangPro \citep{abzianidze-2020-learning} learns lexical relations by
abduction from labelled train/trial examples, then searches automatically at
test time. Our French first-order test result has 54.05\% macro-$F_1$ and
14.95\% commitment coverage.
Natural tableaux decompose typed LLFs, whereas Prover9 resolves enriched
FOL theories, which are obtained with some additional steps after lambda-term generation (lambda-term-to-DRS, then DRS-to-FOL).

We also evaluate the fixed encoder
mDeBERTa-v3$_{\mathrm{Base,NLI\text{-}2mil7}}$
\citep{https://doi.org/10.48550/arxiv.2111.09543}\footnote{The
evaluated checkpoint is
\url{https://huggingface.co/MoritzLaurer/mDeBERTa-v3-base-xnli-multilingual-nli-2mil7}.}
without target-corpus fine-tuning (Table~\ref{eval:table-encoders}). This is
the strongest three-way encoder checkpoint in the comparison across French NLI
corpora, as reported by \citet{skandalis-etal-2024-new}. On GQNLI-FR, both evaluations cover 300 pairs; the reported proof-system
scores exceed the encoder scores by 32.00 accuracy points and 32.06
macro-$F_1$ points. FraCaS shows a smaller difference. On SICK, higher
proof-system accuracy accompanies lower macro-$F_1$: precise selective
deduction has weaker class-balanced performance, consistent with the
neutral-class contribution in Section~\ref{eval:aggregate}.

\begin{table}[!tbp]
\centering
\begin{adjustbox}{max width=\textwidth}
\small
\setlength{\tabcolsep}{5pt}
\begin{tabular}{lrrrcrr}
\hline
 \multirow{2}{*}{Corpus}& \multirow{2}{*}{$n_{\rm enc.}/n_{\rm FOL}$}& \multicolumn{2}{c}{mDeBERTa-v3} &\footnotesize{Best LLM accuracy reported}& \multicolumn{2}{c}{\textbf{FOLLIE + Prover9}}\\
 &  & Acc. & Macro-$F_1$ &\footnotesize{by \citet{Beauchemin-2026}}& Acc. & Macro-$F_1$\\
\hline
FraCaS-FR & 335/335 & 61.49 & 52.95 &66.57 \scriptsize{(Command-r-08-2024)}& 65.37 & 57.47\\
GQNLI-FR & 300/300 & 38.33 & 38.58 &70.00 \scriptsize{(GPT-4.1, Claude-sonnet-4-5)}& 70.33 & 70.64\\
SICK-FR & 9,840/9,399 & 66.94 & 66.51 &83.92 \scriptsize{(o1-2024-12-17 (reasoning))}& 71.35 & 54.36\\
\hline
\end{tabular}
\end{adjustbox}
\caption{French fixed-encoder and first-order results (percent), with their
evaluated population sizes. SICK uses aggregate populations of different
sizes; its row is descriptive rather than paired. Encoder outputs cover all
three labels; the proof engine's unknown label also records unsuccessful
proof search.}
\label{eval:table-encoders}
\end{table}

\subsection{What the experiment establishes}
\label{eval:discussion}

We obtain a precise selective classifier across three French benchmarks,
using a shared theory-construction policy and inspectable proofs \citep{rt_thesis}. Coverage
is the main limitation, particularly contradiction recall in scene
descriptions. Participant identity, aspectual boundaries, quantifier
restrictors and reference populations provide concrete targets for improving
the representations. Controlled axiom ablations and proof-dependency
analysis would separate their effects from those of search limits.

\section{Conclusion and perspectives}
\label{sec:conclusion}

FOLLIE supplies a common experimental basis for French sentence-to-logic
construction and automated inference. Its 36,987 aligned mappings retain
the problem structure of six heterogeneous NLI sources, connecting
French lexical interpretation to the entailments and contradictions that
motivated those sources. The construction method combines learned
grammatical alternatives with explicit composition and a first-order semantic
interface. Stage-wise results show that controlled category ambiguity can
substantially broaden grammatical coverage, while the preserved intermediate
representations make conversion failures localisable.

The resource analysis reveals distinctions that sentence length or a single
formula count cannot capture. Conjunction distributions describe the
accumulation of event, participant and modifier information; operator
incidence and scope alternation distinguish the structural demands
of scene descriptions, quantifier diagnostics and less controlled texts.
Together with source provenance and duplication analysis, these profiles
support selecting and interpreting data for French semantic parsing and
reasoning.

The Prover9/Mace4 study demonstrates how these representations can support
automatic, inspectable NLI. Representation-conditioned lexical, numerical
and structural axioms yield high reference-label agreement among
committed decisions on FraCaS-FR, GQNLI-FR and SICK-FR. The GQNLI-FR
results demonstrate the practical value of explicit quantificational
structure for deduction. Coverage remains uneven: FraCaS reveals
construction-specific dependencies, and SICK highlights the need to preserve
scene identity and license lexical transfer. Proof traces connect these
outcomes to particular representations and assumptions, giving the error
analysis a level of resolution unavailable from an aggregate label score.

The next advances concern the interfaces demonstrated here: independently
assessed sentence interpretations, provenance-preserving resource versions,
scope and reference-set representations, and background theories with
explicit linguistic licensing conditions. Grouped evaluation and controlled
component comparisons can establish how these improvements change the
precision--coverage balance beyond benchmark-informed development.
FOLLIE makes those questions experimentally accessible in French by
bringing sentences, compositional analyses and their inferential
consequences into one resource and reasoning framework.

These findings motivate four connected research directions. At the
grammatical interface, the DeepGrail linker\footnote{\url{https://gitlab.irit.fr/pnria/global-helper/deepgrail-linker}.}
offers an alternative to GrailLight's proof search. Following neural
proof-net approaches \citep{kogkalidis-etal-2020-neural}, it predicts links
between opposite-polarity occurrences of the same atomic type in the supertags.
Supplying, for example, the 128 or 256 most probable category sequences
would test whether learned linking can preserve the coverage benefits of
multitagging while reducing search cost. Candidate linkings must satisfy
the grammar's proof-net correctness conditions before semantic composition.
Holding the semantic lexicon and reasoning engine fixed would allow a
comparison of runtime, derivation coverage, sentence-level semantic fidelity
and downstream inference accuracy.

Contextual lexical interpretation is a second priority. An LLM could assess
whether a relation retrieved from Open Multilingual Wordnet
\citep{bond-foster-2013-linking} or JeuxDeMots
\citep{lafourcade:lirmm-00200883} is licensed by the premises and hypothesis.
For example, the relations \emph{femme isa homme} and
\emph{femme antonym homme} can reflect the human-being and adult-male senses
of \emph{homme}; their applicability depends on the occurrence being
interpreted. A contextual decision could replace coarse exclusions such as
the present blocklist for \texttt{homme}. Alternatively, the LLM could
propose missing lexical bridges directly from the sentence pair, guided by
unresolved proof obligations. Both approaches require checking relation
direction, predicate arguments and the conditions under which an opposition
licenses incompatibility. Accepted axioms should retain their source and
contextual justification; their linguistic validity requires separate
assessment from the correctness of a proof that uses them. Comparing
resource-only, resource-plus-LLM and LLM-proposed theories would establish
their respective effects on precision and coverage.

A third direction, manual gold-standard curation of the entire FOLLIE collection,
would require, given the size and the nature of the dataset, a dedicated funded project and paid annotators trained in
formal semantics and first-order logic. Guidelines should address scope,
negation, coreference, temporal interpretation, predicate senses and argument
structure, with independent annotation and adjudication of disagreements.
Where a sentence permits several defensible readings, the resource should
record the alternatives and their interpretation conditions. Reviewing
sentences within their original inference problems would preserve
cross-sentence reference and vocabulary consistency without forcing formulas
to reproduce an inherited NLI label. Versioned corrections, annotation
provenance and agreement analyses would support both a reliable evaluation
set and subsequent extension of gold coverage to all constituent corpora.

Gold FOLLIE would support training and fine-tuning LLMs for direct
French NL-to-FOL conversion, potentially replacing the linguistic pipeline
with a single learned semantic parser. Comparable reliability remains to be
established.
Evaluation should measure formula well-formedness, binding and argument
consistency, semantic fidelity and downstream inference, using logical
equivalence to gold interpretations where it can be established.
Inconclusive equivalence checks must remain distinct from demonstrated
inequivalence. Splits grouped by repeated sentences and source inference
problems would limit overlap, while targeted tests of scope, negation and
lexical substitution would probe compositional generalisation. Crossing
expert-corrected and predicted formulas with alternative lexical theories
would further distinguish interpretation errors from knowledge gaps.
These directions would clarify when learned interpretation and explicit
deduction jointly improve French language understanding.
\begin{acknowledgments}
The research was funded by the Defence Innovation Agency
(AID--DGA), French Ministry of Armed Forces, and by the Cybersecurity Institute of Occitanie (ICO), to which both we express our gratitude.
\end{acknowledgments}
\newpage
\starttwocolumn
\bibliographystyle{compling}
\bibliography{references}
\end{document}